\documentclass{article}

\usepackage{arxiv}

\usepackage[utf8]{inputenc} % allow utf-8 input
\usepackage[T1]{fontenc}    % use 8-bit T1 fonts
\usepackage{hyperref}       % hyperlinks
\usepackage{url}            % simple URL typesetting
\usepackage{booktabs}       % professional-quality tables
\usepackage{amsfonts}       % blackboard math symbols
\usepackage{nicefrac}       % compact symbols for 1/2, etc.
\usepackage{microtype}      % microtypography
\usepackage{lipsum}		% Can be removed after putting your text content
\usepackage{graphicx}
\usepackage{natbib}
\usepackage{doi}
\usepackage{tikz-cd}
\usepackage{mathtools}

\usepackage{amsmath}
\usepackage{amssymb}
\usepackage{amsthm}
\usepackage[boxruled]{algorithm2e}

\newtheorem{theorem}{Theorem}
\newtheorem{definition}{Definition}
\newcommand{\CI}{\mathrel{\perp\mspace{-10mu}\perp}}

\title{Categoroids: Universal Conditional Independence}

\author{ Sridhar Mahadevan \\
	Adobe Research and University of Massachusetts, Amherst\\
	\texttt{smahadev@adobe.com, mahadeva@umass.edu}
}

\hypersetup{
pdftitle={A template for the arxiv style},
pdfsubject={q-bio.NC, q-bio.QM},
pdfauthor={David S.~Hippocampus, Elias D.~Striatum},
pdfkeywords={First keyword, Second keyword, More},
}

\begin{document}
\maketitle

\begin{abstract}
Conditional independence has been widely used in AI, causal inference, machine learning,  and statistics to represent {\em ternary} geometric, logical, probabilistic, statistical, or topological relationships.  We introduce {\em categoroids}, an algebraic structure for characterizing  universal properties of conditional independence.  Categoroids are defined as a hybrid of two categories: one encoding a preordered lattice structure defined by objects and arrows between them; the second dual parameterization involves {\em trigonoidal} objects and morphisms defining a conditional independence structure, with {\em bridge} morphisms providing the interface between the binary and ternary structures.  A morphism from a trigonoidal object (A, C, B) -- denoting an abstract property that A is conditionally independent of  B given a ``separoid" object C -- to another trigonoidal object (A', C', B') represents a valid proof in a conditional independence axiom system. We illustrate categoroids using three well-known examples of axiom sets: graphoids, integer-valued multisets, and separoids. {\em Functoroids} map  one categoroid to another, preserving the relationships defined by all three types of arrows in the co-domain categoroid. We describe a natural transformation across functoroids, which is natural across regular objects and trigonoidal objects, to construct universal representations of conditional independence. We use a variant of the Yoneda Lemma for categoroids, which defines a natural transformation between the set-valued functoroid induced over all binary, bridge, and trigonoidal arrows from an object to any set-valued functoroid from a categoroid. We use adjunctions and monads between categoroids to abstractly characterize faithfulness of graphical and non-graphical representations of conditional independence. A ``forgetful" functoroid defines a right adjoint between a categoroid and its concrete graphical or non-graphical representation; its associated left adjoint defines the ``free" object associated with categoroid. We study  modular Gr\"obner representations of categoroids and identify the conditions under which they are noetherian, that is, they satisfy the ascending chain condition for partially ordered sets, thus giving rise to a Gr\"obner basis. 
\end{abstract}

% keywords can be removed

\section{Introduction}

\begin{figure}[h] 
\centering
\caption{Categoroids and their relationships. To enhance readability, not all links are shown. \label{categoroids}}
\vskip 0.1in
\begin{minipage}{0.9\textwidth}
\includegraphics[scale=0.4]{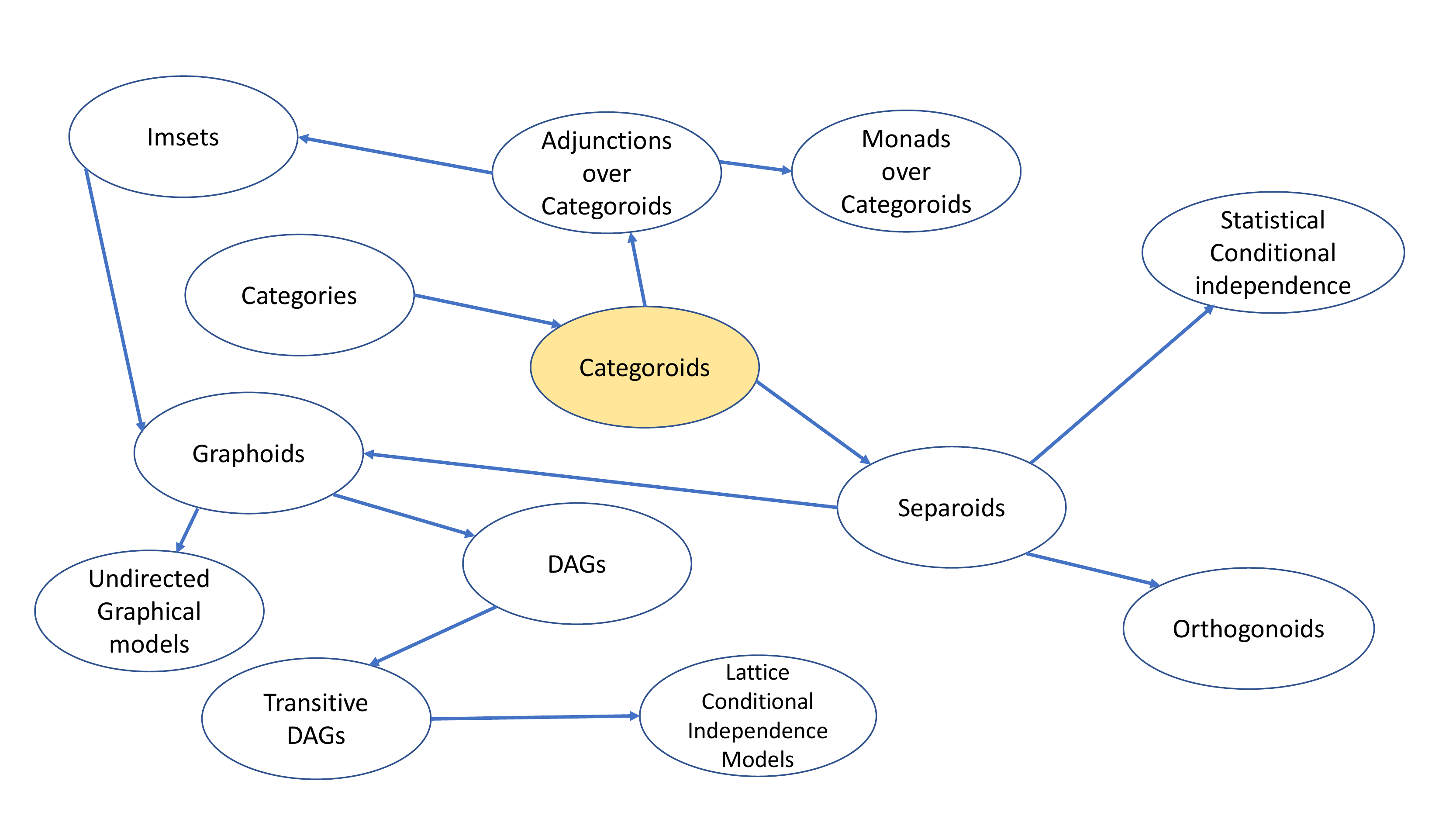}
\end{minipage}
\vskip 0.1in
 \end{figure}

A {\em categoroid} is  defined by a ``flexible" join of two categories, one over the usual objects and binary arrows, and the other over trigonoidal objects, or triples, and their associated morphisms. \footnote{Recall that a category ${\cal C}$ is defined by a collection of objects ${\cal O}$, and a collection of morphisms $f: a \rightarrow b$, where $a,b \in {\cal O}$, each object $c$ is equipped with an identity morphism {\bf 1}$_c$, and morphisms satisfy associative and compositional properties \citep{maclane:71}.} Bridge arrows define the join between the two categories.  Categoroids are used in this paper to analyze universal properties underlying conditional independence, illustrated by previous axiomatizations such as {\em graphoids} \citep{pearl:bnets-book,pearl:causalitybook}, {\em integer-valued multisets} \citep{studeny2010probabilistic,matus}, and {\em separoids} \citep{DBLP:journals/amai/Dawid01,DBLP:journals/jmlr/Dawid10}. Categoroids are a type of category \citep{maclane:71}, with a novel dual parameterization, one involving the usual unary collection of objects and binary arrows mapping between them, but the other comprised of a collection of trigonoidal morphisms that act over triples of objects, with bridge morphisms defining the join. The trigonoidal morphisms represent a proof system defined over a conditional axiom system. Any valid conditional independence property can be inferred as a sequence of trigonoidal morphisms if such a property can be inferred by using the axiom system. Our primary interest is in developing a deeper understanding of how these two parameterizations interact. Figure~\ref{categoroids} gives an overview of the different categoroids that will be discussed in this paper. The arrow indicates each  categoroid in the co-domain is a special case of the categoroid in the domain. Not all relationships are shown to enhance readability.

Conditional independence structures have been actively studied in AI, causal inference, machine learning, probability, and statistics for many years. \citet{DBLP:journals/amai/Dawid01,DBLP:journals/jmlr/Dawid10} define separoids, a join semi-lattice, to formalize reasoning about conditional independence and irrelevance in many areas, including statistics. \citet{pearl:bnets-book} introduced {\em graphoids}, a distributive lattice over disjoint subsets of variables, to model reasoning about irrelevance in probabilistic systems, and proposed  representations using directed acyclic graphs (DAGs).  \citet{studeny2010probabilistic} proposed a lattice-theoretic model of conditional independences using integer-valued multisets to address the intrinsic limitations of DAG-based representations. Designing suitable graphical and non-graphical representations of conditional independence structures continues to be actively studied in the literature \citep{lauritzen:chain,mdag,hedge,amini,cif,sadeghi}. 

To illustrate the main themes of this paper, consider a directed acyclic graph (DAG) $G = (V, E)$ as a representation of a conditional independence relation set, where the  conditional independence $\CI$ property is defined using the graph property of $d$-separation \citep{pearl:bnets-book}. A given DAG $G$ can be characterized in two ways: one parameterization specifies the DAG $G$ in terms of the vertices $V$  and  edges $E$, which corresponds to specifying the objects and morphisms of a categoroid. The second way to parameterize a DAG is by its induced collection of conditional independence properties, as defined by $d$-separation. For example, the serial DAG over three variables, $A \rightarrow B \rightarrow C$, can be defined using its two edges $A \rightarrow B$ and $B \rightarrow C$, but also by its conditional independences, namely $A \CI C | B$ using the theory of $d$-separation. We are thus given two possibly redundant parameterizations of the same algebraic structure. However, multiple DAG models can define the same conditional independences. For example, the serial model $A \rightarrow B \rightarrow C$, as well as the ``diverging" model $A \leftarrow B \rightarrow C$ and the ``reverse" serial model $A \leftarrow B \leftarrow C$ all capture the same conditional independence property $(A \CI C | B)$. \footnote{This non-uniqueness property arises because Bayes rule can be used to reparameterize any one of these three DAGs into the form represented by one of the other DAGs.}  

Categoroids are more general than previous graphical and non-graphical representations. A separoid $({\cal S}, \leq, \CI)$ \citep{DBLP:journals/amai/Dawid01} defines a semi-lattice ${\cal S}$, where the join $\vee$ operator over the semi-lattice ${\cal S}$ defines a preorder $\leq$, and the ternary relation $\CI$ is defined over triples of the form $(x \CI y | z)$ (which are interpreted to mean $x$ is conditionally independent of $y$ given $z$).  We can define a ``separoid" induced categoroid ${\cal S}$ as a category, where the morphism $f: x \rightarrow y$ defines a preorder $x \leq y$. As pointed out by \citet{maclane:71}, ``lattice" objects can be constructed in any category. It is then possible to define an abstract ``separoid" ternary relation in any category, because each morphism $f: x \rightarrow y$ yields a {\em bridge} morphism $b: (x,y) \rightarrow (x, y, x)$ (over all objects $z$ in {\cal C}) (abstracting from the property in separoids that $x \leq y$ implies that $(x \CI x | \ y)$).  \footnote{A notational remark: we often write $(x,z,y)$ to denote the conditional independence $(x \CI y | z)$, when we wish to abstract from the somewhat loaded semantics associated with $\CI$ in the previous literature,  similar to the graphoid notation of \citet{pearl:bnets-book}, where inserting the conditioning element $z$ in the middle is suggestive of the ``separoid" action as defined by  by \citep{DBLP:journals/amai/Dawid01}.} Furthermore, a trigonoidal morphism can be defined $t: (x, z, y) \rightarrow (x, z, w)$ for any $g: w \rightarrow y$ (abstracting again from the separoid axiom that if $(x \CI y |\ z)$ and $w \leq y$, then it follows that $(x \CI w | \ z)$ for any $x,y,z \in {\cal S}$). Other conditional independence properties in a separoid, for example the symmetry property stating from $(x \CI y | z)$, we can infer $(y \CI x |z)$ can in turn be abstracted into a trigonoidal morphism $t: (x, z, y) \rightarrow (y, z, x)$. 

{\em Functoroids} map from one categoroid ${\cal C}$ to another categoroid ${\cal D}$, mapping the object $o$ in ${\cal C}$ to $F(o)$, an object in ${\cal D}$, arrows $x \rightarrow y$ in ${\cal C}$ to corresponding arrows $F(x) \rightarrow F(y)$ in ${\cal D}$, but uniquely, acting trigonoidically as well, mapping trigonoids $(x,z,y) \rightarrow (x', y', z')$ in ${\cal C}$ to corresponding trigonoids $(F(x), F(y), F(z)) \rightarrow (F(x'), F(y'), F(z'))$ in ${\cal D}$ (and bridge arrows as well).  To explore the mapping between a categoroid and its graphical or non-graphical representation, we use the theory of adjoint functors \citep{maclane:71,richter2020categories,riehl2017category}. We can illustrate this framework using integer-valued sets (or {\em imsets}) \citep{studeny2010probabilistic}, which are another non-graphical approach to representing conditional independence. An imset is defined as an integer-valued multiset function $u: \mathbb{Z}^{{\cal P(\mathbb{Z})}} \rightarrow \mathbb{Z}$ from the power set of integers, ${\cal P(\mathbb{Z})}$ to integers $\mathbb{Z}$. An imset is defined over a distributive lattice of disjoint (or non-disjoint) subsets of variables $A$, $B$, and $C$, over the universe of all variables $N$.  A {\em combinatorial} imset is defined as: \footnote{A {\em structural} imset is defined as one where the coefficients can be rational numbers. Our analysis of imsets using  adjoint functors can be extended to this case as well.}

\[ u = \sum_{A \subset N} c_A \delta_A \]

where $c_A$ is an integer, $\delta_A$ is the characteristic function for subset $A$, and $A$ potentially ranges over all subsets of $N$. An {\em elementary} imset is defined over $(a,b \CI A)$, where $a,b$ are singletons, and $A \subset N \setminus \{a,b\}$. \citet{bouckaert} explore linear programming methods for  conditional independence inferences using imsets. For a general DAG model $G = (V, E)$, an imset in standard form \citep{studeny2010probabilistic} is defined as 

\[ u_G = \delta_V - \delta_\emptyset + \sum_{i \in V} (\delta_{\mbox{{\bf  Pa}}_i} - \delta_{i \cup \mbox{{\bf Pa}}_i}) \] 

where {\bf Pa}$_i$ define the parents of variable $i$ in the DAG. The three canonical DAGs over the variables $G = \{a, b, c \}$, namely the serial DAG $a \rightarrow b \rightarrow c$, the reverse serial DAG $a \leftarrow b \leftarrow c$, and the diverger DAG $a \leftarrow b \rightarrow c$, all yield the same imset representation (see Figure~\ref{imset}): 

\[ u_G = \delta_{abc} - \delta_{ab} - \delta_{bc} + \delta_b \] 

We define adjunctions between functoroids to explore the mapping between the ``free" objects associated with a categoroid, and a ``forgetful" functoroid that maps a given graphical or non-graphical representation back to a categoroid. The family of adjoint functors defined by such ``free" $+$ ``forgetful" functors forms an important aspect of many constructions in category theory \citep{maclane:71,riehl2017category,richter2020categories}. Viewed in this light, imsets and other graphical representations of conditional independence structures emerge as illustrative examples of adjoint functoroids between categoroids. Similarly, we can study graphoids and the entire panoply of graph-based representations, including DAGs \citep{pearl:bnets-book}, marginalized DAGs \citep{mdag}, hyperedge-directed graphs (HEDGes) \citep{hedge}, chain graphs \citep{lauritzen:chain}, and mixed graphs \citep{sadeghi} in terms of the theory of adjoint functoroids between categoroids. For example, we can define a ``forgetful" functoroid from a DAG model to a universal categoroid representing this conditional independence property. Using adjunctions on categoroids, we can view the left adjoint of the``forgetful" functoroid as mapping from the ``free" object in the conditional independence model that generates  all three DAG structures associated with it. 

To explain in a bit more detail the concept of adjunctions between free and forgetful functors, we can analyze imsets in terms of adjunctions defined by a pair of functors between a categoroid and the category of left-{\bf k} modules over a commutative ring {\bf k}[S]  \citep{riehl2017category}. The left-{\bf k} module over a commutative ring defined by a set $S$ is defined formally as 

\[ \mbox{{\bf k}}[S] = \sum_{x \in S} z_x x \] 

where $z_x$ are elements of the ring {\bf k} (which, as in the case of imsets, can be the integers $\mathbb{Z}$ or rationals $\mathbb{Q}$).  A ``forgetful" functor $U$ defines the {\em right-adjoint} between a left {\bf k}-module {\bf k}[S] and the set $S$ (where the Abelian group structure of the commutative ring is ``thrown away" in the category  {\bf Sets}, of which $S$ is an object). The ``free" functor $L$ maps the set $S$ back to the left-{\bf k} module {\bf k}[S]. We generalize the imset construction, developing a more general theory of left-{\bf k} module representations of any categoroid ${\cal C}$, building on recent advances in mathematics on Gr\"obner representations of combinatorial categories \citet{Sam_2016}, as explained below. 

A categoroid ${\cal C}$ is defined by a unary collection of objects $O$, a collection of binary arrows defined as a subset of the binary relation $O \times O$, but crucially and perhaps uniquely  also includes a collection of ternary structures called {\em trigonoids} defined over triples of objects $O \times O \times O$, and their associated morphisms representing the inference of new conditional independence properties.  Trigonoidal objects and their morphisms act as the carrier of  conditional independence in a categoroid. The axiomatic inference of new conditional inference properties has been extensively explored in the previous work on graphoids, imsets, and separoids, among others, but perhaps unique to our paper is the conceptualization of this process as a type of proof theory in a categoroid involving the composition of trigonoidal morphisms. Category theory has long been understood as a way to conceptualize inference rules in logic, where assertions are objects, and proofs are morphisms \citep{Baez_2010}.  As an example, a separoid $({\cal S}, \leq)$ is a semi-lattice with a {\em join} $\vee$ operator, augmented with a ternary relation $(. \ \CI \ . | \ .)$ defined over triples of objects. The term $x \CI y |z$ is typically read as $x$ is conditionally independent of $y$ given $z$. A common use of conditional independence is to {\em gate} the transmission of information from an object $x$ to an object $y$ by conditioning on a third object $z$, which eliminates all paths that carry information from $x$ to $y$. Similarly, if we interpret conditional independence as a statement about ``irrelevance", then we might assert that if $(x \CI y | z) \wedge w \leq y \Rightarrow (x \CI w | z)$. This assertion can be interpreted as stating that if ``knowing" $z$, $y$ is irrelevant to $x$, then we can also surmise that any ``weaker" $w \leq y$ is also irrelevant to $x$ when $z$ is known. The properties of separoids can be viewed as morphisms that map a triple of objects, for example $(x \CI y | x)$ into a new triple $(y \CI x | x)$, reflecting a symmetry principle. A fundamental aspect of conditional independence structures is that they all combine an associative algebraic structure with a ternary relation over triples of objects. Trigonoids formulates these ternary relation using the actions of trigonoidal morphisms. Unlike previous approaches, categoroids make no assumptions regarding the finiteness of conditional independence axiomatizations. A collection of trigonoids need not be finite, and can capture non-finite axiomatizations \citep{studeny2010probabilistic}.

Generalizing from previous axiomatizations of conditional independence, in a categoroid, we interpret an arrow $f: x \rightarrow y$ as defining a (non-unique) preorder relation $x \leq y$. We can conceptualize a categoroid as inducing two subcategories, one modeling the regular morphisms, and the other representing the trigonoidal morphisms over triples of objects. However, these structures interact in interesting ways, and while categoroids can be viewed as a join of two categories, their interaction is what makes categoroids (and previous axiomatizations) unique. We can abstractly define a ``forgetful" functor that converts a categoroid into a category by discarding one of the two parameterizations, which would correspond to projecting a separoid $({\cal S}, \leq, \CI)$ on one of its two parameterizations. The collection of morphisms out of an object $x$ in a category ${\cal C}$ is typically denoted as {\bf Hom}$_{\cal C}(x, -)$, but also more succinctly as ${\cal C}(x, -)$, constitutes a set-valued functor called a {\em co-presheaf} \citep{maclane:sheaves}. We include in the presheaf all bridge and trigonoidal morphisms``out" of a trigonoidal object containing $x$.  For each morphism $f: x \rightarrow y$, we define a collection of trigonoidal morphisms that are induced by a given axiomatization of conditional independence. For example, the axioms of separoids permit inferring from $x \leq y$, the conditional independence $x \CI x | y$.  Using a further property of conditional independence, an entire (possibly non-finite) collection of trigonoidal morphisms can be defined from the regular morphism $f: x \rightarrow y$ as $t_w: (x, y, x) \rightarrow (x, y, w)$ for all $g: w \rightarrow x$. This process generalizes the case in separoids where from $x \CI x | y$ and $w \leq x$, it is possible to infer $x \CI w | y$. Similarly, a ``basic" separoid $({\cal S}, \leq, \CI)$ is defined as one where $x \CI x | y \ \Rightarrow x \leq y$, leading to a notion of a ``basic" categoroid as one where a trigonoidal object $(x, y. x)$ can be used to define a morphism $x \rightarrow y$. In a DAG model, the co-presheaf contains all the directed paths entering a vertex $x$ in a DAG. 

Using the property that imsets are essentially functions on the distributive lattice of sets, which define partially ordered sets (posets), we use the theory of M\"obius inversions \citep{mobius} to decompose each imset into a convolution of the {\em zeta} function $\zeta(x,y)$ (which is equal to $1$ for all $x \leq y$ in a poset with finite intervals), and the m\"obius function $\mu(x,y)  = -\sum_{x \leq z < y} -\mu(x,y)$, which reveals the connection between imsets and categoroids, showing how each imset can be expressed in terms of the arrow structure of a categoroid. In particular, for the imset representation, the presheaf is defined by the m\"obius inversion of an imset, which represents the imset in terms of a function defined over all morphisms entering the lattice element defined by the subset $S \subset V$ of all variables. Thus, an important theme of our paper is that the use of category-theoretic notions, such as adjunctions, functors, and presheafs provides an abstract language to analyze previous axiom systems of conditional independence. 

It is possible to combine categoroids in interesting ways. Given a collection of separoid relations $\{\CI_\alpha | \alpha \in a \}$, generalizing the construction given in \citep{DBLP:journals/amai/Dawid01}, we can define an induced categoroid over a collection of trigonoidal relations, each defined by a collection of trigonoidal morphisms. In the case of separoids, each $\CI$ relation is defined as a set of triples $(x,y,z) \in S \times S \times S$, and the set of all such $\CI$ relations forms a lattice itself, where two $\CI$ relations can be combined using joins (unions) or meets (intersections). Generalizing from these ideas in separoids, we can define corresponding structures in categoroids.  As \citet{maclane:71} pointed out, it is easy to construct ``lattice" objects in any category in terms of the arrows $x \rightarrow y$. Thus, we can construct ``separoid"-like structures in any category as well, which generates a rich variety of possible categoroids. For example, we can define {\em topoids} as categoroids that satisfy the axioms of a topos, namely they have all finite colimits and limits, have exponential objects, and a subobject classifier, but additionally admit a dual parameterization in terms of trigonoidal objects and their morphisms. To this end, we prove a novel variant of the Yoneda Lemma, which provides a faithful and full embedding of any categoroid in the category of sets. We construct the Yoneda embedding {\bf Hom}$_{\cal C}(-, x)$ over all types of morphisms  defining the functoroid category of co-presheaves on a categoroid. Co-presheaves over a categoroid define a topoid, and the density theorem for sheaves \citep{maclane:sheaves} extends to cateogoroids as well, allowing a way for constructing abstract diagrams for categoroids. Specifically, abstract categoroid diagrams can be seen as functoroids from some finite indexing category ${\cal J}$ of diagrams. The density theorem for categoroids shows that every presheaf in a categoroid can be represented in a canonical way as the co-limit of a set of representable presheaves. This result has many applications, including Universal Causality \citep{sm:uc}. 

Many other examples of categoroid construction are possible, such as braided categoroids \citep{JOYAL1996164}, and symmetric monoidal categoroids, which has applications in modeling complex compositional systems \citet{fong2018seven}.  \citet{DBLP:journals/amai/AnderssonMPT97} study lattice conditional independence models, defined over a ring of subsets of a set of discrete elements. These models arise naturally in investigating missing data generated from a multivariate normal distribution, where the pattern of missingness forms a lattice. These models are also equivalent to transitive DAG models, which naturally can be viewed as a category. These models can also be parameterized using the order ideals of a distributive lattice. A classic theorem by Hibi \citep{hibi1987distributive} shows that a reduced Gr\"obner basis can be defined over ideals of a poset, which can be applied to construct a toric variety over graphical models, such as Conjunctive Bayesian Networks \citep{cbn}, log-linear statistical models and Markov random fields \citep{Geiger_2006}. This approach can be generalized by defining modular representations of categoroids based on recent work on Gr\"obner categories \citep{Sam_2016}. The key idea underlying this approach is to identify conditions under which a modular representation is {\em noetherian}, that is, it satisfies an ascending chain condition on a partially ordered set, and thereby can give rise to Gr\"obner bases. At a high level, we are given a set-valued functor $S: {\cal C} \rightarrow {\bf Set}$, which maps a categoroid into the category of sets. Each object $x$ is therefore mapped to a set $S(x)$. Using  the representation theory of left {\bf k}-modules over a ring, a standard approach in mathematics for studying many algebraic structures, we can define the left {\bf k}-module $P = {\bf k}[S]$ where $P(x)$ is the free {\bf k}-module on the set $S(x)$. This approach can be seen as a generalization of integer-valued multisets \citep{studeny2010probabilistic}, which as described above has been used to study conditional independence structures.

\section{Categoroids} 

We define categoroids in this section, and relate them to previous axiomatizations of conditional independence, such as graphoids \citep{pearl:bnets-book}, imsets \citep{studeny2010probabilistic}, and separoids \citep{DBLP:journals/amai/Dawid01}. Categoroids build on the formalism of category theory \citep{maclane:71}. We will introduce the terminology as appropriate. A simple way to understand a category is as a {\em quiver}, a directed graph with many (possibly infinite) numbers of directed edges between any two objects representing arrows (or also referred to as morphisms). It is possible to interpret any directed graph as a ``free" category, where the vertices define the objects, and the set of all (finite or non-finite) paths between a pair of vertices are defined as its morphisms.  Succinctly, we can define a category ${\cal C}$ as a collection of objects ${\cal O}$ and arrows {\cal A} with two functions, $\partial_0: {\cal A} \rightarrow {\cal O}$ and $\partial_1: {\cal A} \rightarrow {\cal O}$, which define the ``head" and ``tail" of each arrow. The arrows compose in the usual manner, that is $g \circ f: a \rightarrow c$ is the composition of $f: a \rightarrow b$ and $g: b \rightarrow c$. Each object is associated with a distinguished self loop called the identity arrow {\bf 1}$_c$. The category with one object and one morphism (identity) is denoted ${\bf 1}$. An {\em initial} object in a category is one that defines exactly one morphism between it and any other object in the category (including itself). A {\em terminal object} in a category is one that has exactly one morphism from any object to itself.  

As much of our discussion below will involve lattices, it is useful to relate these definitions to lattices and partially ordered sets. Each element in the lattice {\cal L} defines an object in the category. The morphisms in the category are defined by the preorder (partial) order associated with the (semi)lattice (or poset). Any category induces a (reflexive, transitive) preorder $\leq$ by defining $o \leq o'$ if there is a morphism $f: o \rightarrow o'$. The initial element in the lattice is the bottom element, often denoted $\perp$ (or $\emptyset$ for lattices defined over subsets). The final element in a lattice is denoted as $\top$ (or the entire universe of objects, e.g., the set of vertices $V$ in a graphical model). 

Given two objects in a category, there can be a non-finite number of arrows between them. For example, consider the category {\bf Top} of all topological spaces. Each object $o = \{X, {\cal T} \}$ is a set $X$, and a collection of open sets ${\cal T}$ closed under finite intersection and arbitrary unions. An arrow $f: o \rightarrow o'$ between two topological spaces is a continuous function that maps each element $x \in X$ to a corresponding element $f(x) \in X'$, where $o' = \{X', {\cal T}' \}$, such that the preimage $f^{-1}(A)$ of any open set $A \subset {\cal T}'$ is an open set in ${\cal T}$, that is $f^{-1}(A) \subset {\cal T}$. In general, there can be a non-finite number of such continuous functions between any two topological spaces. A category is called {\em locally finite} if there exists only a set of arrows between any two objects (note the set may not be finite). In our paper, we assume all categories are locally finite. In addition, for some of the results, it may be useful to restrict the category to a finite number of objects, and a finite number of morphisms between a pair of objects, such as in the application to graphical models over a finite number of variables. For simplicity, we will endeavor to keep the category-theoretic formalism at an elementary level. The definition of categoroids proposed below can be enhanced using fancier constructions, such as those given by \citet{fong2018seven}, including braided or symmetric monoidal categories, decorated cospans, and operads. We leave these embellishments to a future paper. 

\begin{definition}
\label{cat-defn}
A {\bf conditional independence quiver} is a set $O$ of {\em objects}, a set $A$ of {\em arrows}, a set $T$ of {\em trigonoidal arrows}, and two sets $B^0,B^1$ of {\em bridge arrows}, along with pairs  of functions $\partial^A_0: A \rightarrow O$ and $\partial^A_1: A \rightarrow O$, $\partial^T_0: T \rightarrow O \times O \times O$, $\partial^T_1: T \rightarrow O \times O \times O$, $\partial^{B^0}_0: B^0 \rightarrow O \times O$,   $\partial^{B^{0}}_1: B^0 \rightarrow O \times O \times O$, and finally, $\partial^{B^1}_0: B^1 \rightarrow O \times O \times O$,  and  $\partial^{B^{1}}_1: B^1 \rightarrow O \times O$ specifying the domain (``head") and co-domain (``tail") of each type of arrow, respectively. \footnote{We are not assuming that conditional independence quivers are finite, unless explicitly stated in the paper. it is possible to have a (an infinite) set of arrows of any type between the appropriate object types.}. We define the composable pairs of arrows and trigonoids as the sets:

\begin{eqnarray}
A \times_0 A = \{\langle g, f \rangle | \ g, f \in A \ \ \mbox{and} \ \ \partial^A_0(g) = \partial^A_1(f) \} \\
T \times_0 T = \{\langle g, f \rangle | \ g, f \in T \ \ \mbox{and} \ \ \partial^T_0(g) = \partial^T_1(f) \} \\
B^1 \times_0 B^0 = \{\langle g, f \rangle | \ g \in B^1, f \in B^0 \ \ \mbox{and} \ \ \partial^{B^1}_0(g) = \partial^{B^0}_1(f) \} \\
B^0 \times_0 B^1 = \{\langle g, f \rangle | \ g \in B^0, f \in B^1 \ \ \mbox{and} \ \ \partial^{B^0}_0(g) = \partial^{B^1}_1(f) \}
\end{eqnarray}

or in words, the composition $g \circ f$ is possible when the ``domain" of $g$ is equal to the ``range" (or co-domain) of $f$ for all types of arrows. Note in particular that $B^0$ arrows can only combine with $B^1$ arrows, and vice versa, $B^1$ arrows can only combine with $B^0$ arrows. A {\bf categoroid} is the free category over the conditional independence quiver with these additional functions, under the quotient equivalence class defined by ignoring the distinction between primitive morphisms and compositions of morphisms:

\begin{eqnarray}
O \xrightarrow[]{\mbox{id}} A \\
A \times_0 A \xrightarrow[]{\circ} A \\
T \times_0 T \xrightarrow[]{\circ} T \\
B^1 \times_0 B^0  \xrightarrow[]{\circ} B^0 \\
B^0 \times_0 B^1  \xrightarrow[]{\circ} B^1
\end{eqnarray}

\end{definition}

\citet{lurie} defines the join of two categories ${\cal C}$ and ${\cal D}$ as one where the objects are defined as the disjoint unions of the objects of each, and the morphisms $f: a \rightarrow b$ correspond to the morphisms of ${\cal C}$ if $a, b$ are from ${\cal C}$, whereas if $a, b$ are from ${\cal D}$, the morphism $f$ is defined as in ${\cal D}$. He defines the join over pairs $(a,b)$ of objects across the two categories differently than we do above. In our case, the bridge morphisms serve as the interface between the two categories whose join defines a categoroid. 

\subsection{Separoids, Integer-valued Multisets and Graphoids} 

We illustrate the definition of categoroids above using three well-known axiomatizations of conditional independence: separoids \citep{,DBLP:journals/amai/Dawid01}, graphoids \citep{pearl:bnets-book}, and imsets \citep{studeny2010probabilistic}. 

\subsubsection{Separoids are Categoroids over Lattices}

\begin{theorem}
\label{separoid}
A {\bf separoid} \citet{DBLP:journals/amai/Dawid01} defines a categoroid over preordered set $({\cal S}, \leq)$, namely $\leq$ is reflexive and transitive, equipped with a {\em ternary} relation $\CI$ on triples $(x,y,z)$, where $x, y, z \in {\cal S}$ satisfy the following properties: 
\begin{itemize}
    \item {\bf S1:} $({\cal S}, \leq)$ is a join semi-lattice. 
    \item {\bf P1:} $x \CI y \ | \ x$
    \item {\bf P2:} $x \CI y \ | \ z \ \ \ \Rightarrow \ \ \ y \CI x \ | z$ 
    \item {\bf P3:} $x \CI  y \ | \ z \ \ \ \mbox{and} \ \ \ w \leq y \ \ \ \Rightarrow \ \ \ x \CI w \ | z$ 
    \item {\bf P4:} $x \CI y \ | \ z \ \ \  \mbox{and} \ \ \ w \leq y \ \ \ \Rightarrow \ \ \ x \CI y \ | \ (z \vee w)$
    \item {\bf P5:} $x \CI y \ | \ z \ \ \ \mbox{and} \ \ \ x \CI w \ | \ (y \vee z) \ \ \ \Rightarrow \ \ \ x \CI (y \vee w) \ | \ z$
\end{itemize}

A {\bf strong separoid} also defines a categoroid. A strong separoid is defined over a lattice ${\cal S}$ has in addition to a join $\vee$, a meet $\wedge$ operation, and satisfies an additional axiom: 

\begin{itemize} 

\item {\bf P6}: If $z \leq y$ and $w \leq y$, then $x \CI y \ | \ z \ \ \ \mbox{and} \ \ \ x \CI y \ | \ w \ \ \ \Rightarrow \ \ \ x \CI y  \ | \ z \ \wedge  \ w$
\end{itemize}

\end{theorem} 

{\bf Proof:} The proof is quite straightforward, and involves showing how each of the elements in a separoid can be represented by an element of a categoroid. 

\begin{itemize} 

\item The set $O$ of objects in a separoid is simply the elements of the preordered set ${\cal S}$. 

\item The set $A$ of arrows is defined as the morphisms $f: x \rightarrow y$ if $x \leq y$ in the preorder. 

\item The set of trigonoidal morphisms $T$ is defined using the properties $P1$ through $P6$ above. For example, property $P2$ induces a trigonoidal morphism $(x, z, y) \rightarrow (y, z, x)$. \footnote{As a reminder, we are using the notation introduced by \citet{pearl:bnets-book} and placing the conditioning element $z$ in the middle to indicate its action as a ``separoid"  between $x$ and $y$.} 

\item The set of bridge arrows $B^0$ act as a connection between the preorder relation $x \leq y$ to the conditional independence statement $x \CI x | y$.  Stated more generally, each bridge arrow here is written as $f: (x, y) \rightarrow (x, y, x)$ precisely when $f: x \rightarrow y$. 

\item Finally, the set of bridge arrows $B^1$ act as a connection between the conditional independence statement $x \CI x | y$ to the preorder $x \leq y$. Stated in more general terms, $g: (x, y, x) \rightarrow (x,y)$  

\end{itemize}  

To explain the construction above in more detail, our goal is to ensure that every inference of a property in a separoid is captured by an arrow of some type, from a regular arrow $f: x \rightarrow y$ corresponding to the preorder relation $x \leq y$, to the trigonoidal arrow $f: (x, z, y) \rightarrow (y, z, x) $ capturing the symmetry property $x \CI y | z \ \ \rightarrow y \CI x | z$, and finally to the two bridge arrows connecting the preorder $\leq$ relation with the ternary $\CI$ relation. To show that categoroids are more general than separoids, note that we can generalize the join $\vee$ operator in a semi-lattice to any coproduct in a categoroid, or even more generally, to a colimit or even a Kan extension \citep{maclane:71}. Coproducts, colimits, and Kan extensions are universal constructions in a category that assemble into a unified structure a diverse set of similar constructions. For example, the join operator in separoids is the same as the union of two (disjoint) sets in graphoids and imsets, and all of these define coproducts or colimits in a categoroid. The following commutative diagram captures the universal property underlying coproducts. The below figure shows a diagram, a standard construct in category theory, where objects are depicted by vertices with labels, and morphisms are indicated by labeled edges. 

\begin{center}
\begin{tikzcd}
%  T
%  \arrow[drr, bend left, "x"]
%  \arrow[ddr, bend right, "y"]
%  \arrow[dr, dotted, "r" description] & & \\
    & Z\arrow[r, "p"] \arrow[d, "q"]
      & X \arrow[d, "f"] \arrow[ddr, bend left, "h"]\\
& Y \arrow[r, "g"] \arrow[drr, bend right, "i"] &X \sqcup Y \arrow[dr, "r"]  \\ 
& & & R 
\end{tikzcd}
\end{center} 

In the commutative diagram above, the coproduct object $X \sqcup Y$ uniquely factorizes any arrow $h: X \rightarrow R$ and any arrow $i: Y \rightarrow R$, so that $h = r \circ f$, and furthermore $i = r \circ g$. Coproducts are themselves special cases of the more general notion of colimits which will be defined below. The object $X \sqcup Y$ is called a {\em universal element} \citep{riehl2017category} because it represents the universal property of coproducts. Thus, defining a categoroid using coproducts generalizes the use of joins in separoids. The separoid axioms can then be generalized as well, replacing each occurrence of the join $\vee$ above with the coproduct $\sqcup$ operator.  $\bullet$

\subsection{Graphoids are Categoroids over Graphs} 

\citet{pearl:bnets-book} introduced an axiomatization of irrelevance to study causal and probabilistic reasoning over graphical models called {\em graphoids}.  \citep{DBLP:journals/amai/Dawid01} shows that graphoids can be characterized as strong separoids on a distributive lattice having a minimal element $\emptyset$ and possessing relative complements, with the added condition that the join $\vee$ operation is defined on pairwise disjoint terms. We show more directly that semi-graphoids and graphoids define categoroids over a ring of disjoint subsets, where joins are defined by unions and meets are defined by intersections. 

\begin{theorem}
\label{graphoid} 
Semi-graphoids and graphoids define categoroids. A {\em graphoid} {\cal G} is defined over a universe $U$ of (typically discrete) variables, where given any three disjoint subsets of variables $X$, $Y$ and $Z$, the graphoid satisfies the following ternary relationship specified by $I(X,Z,Y)$ (which is to interpreted as $X$ is {\bf independent} of $Y$ given $Z$): 

\begin{itemize}
\item {\bf Symmetry:} $I(X, Z, Y) \ \ \ \Longleftrightarrow \ \ \ I(Y, Z, X)$

\item {\bf Decomposition:} $I(X, Z, Y \cup W) \ \ \Rightarrow \ \ \ I(X, Z, Y) \wedge I(X, Z, W)$

\item {\bf Weak Union:} $I(X, Z, Y \cup W) \ \ \Rightarrow \ \ \ I(X, Z \cup W, Y)$

\item {\bf Contraction: } $I(X, Z, Y) \wedge I(X, Z \cup Y, W) \ \ \Rightarrow \ \ I(X, Z, Y \cup W)$

\item {\bf Intersection:} For strictly positive distributions $P$ whose independence properties are being captured by $I$, the following additional property holds as well: $I(X, Z \cup W, Y) \wedge I(X, Z \cup Y, W)  \ \ \Rightarrow I(X, Z, Y \cup W)$
\end{itemize} 
\end{theorem} 

 {\bf Proof:} The proof follows along the lines of the one given above for separoids. We note that the objects of the categoroid are defined as all subsets of a finite collection of variables. The binary arrows are defined using containment relationships between subsets. Each graphoid property above defines a trigonoidal morphism, following the approach used above for separoids. Similarly, the bridge morphisms can also be defined analogously as was done for separoids. It is worth explicitly noting how the trigonoids defined over these properties capture conditional independence over probability distributions. Specifically, if $I(X, \emptyset, Y \cup W)$ holds, this is tantamount to asserting that 

\[ P(X = x, Y = y, W = w) = P(X = x) P(Y = y, W = w) \ \ \ \forall x, y, w\]

That is, the joint distribution $P$ over $X,Y,W$ factors into a product of two smaller distributions as specified above. \citet{pearl:bnets-book} explores many variants of these axioms for specific graph structures, for example causal DAG models satisfy the following additional axioms: 

\begin{itemize}
    \item {\bf Weak Transitivity:} $I(X, Z, Y) \wedge I(X, Z \cup \gamma, Y) \ \ \Rightarrow \ \ I(X, Z, \gamma) \vee I(\gamma, Z, Y) $
    
    \item {\bf Chordality:} $I(\alpha, \gamma \cup \delta, \beta) \wedge I(\gamma, \gamma, \beta) \ \ \vee \ \ I(\alpha, \delta, \beta)$
\end{itemize}

Here, $\alpha, \beta, \gamma$ denote individual variables in the model. All of these variants can be easily represented as morphisms over trigonoidal objects in a categoroid. \citep{fong:ms} shows how Bayesian networks and DAG models can be represented using symmetric monoidal categories, where the parents  {\bf Pa}$_c$ of a variable $c$, namely $a,b$, in a collider DAG $a \rightarrow b \leftarrow c$ can be ``tensored" together as the composite object $a \otimes b$ using the symmetric monoidal structure of the category. Such symmetric monoidal categoroids can in turn be used to define symmetric monoidal categoroids, by including as well trigonoidal objects and morphisms capturing graphoid properties.  Importantly, a wider class of graphical models can be defined using universal constructions. For example, a universal collider  $a \xrightarrow[]{f} c \xleftarrow[]{g} b$ is defined as the universal construction of a pullback in category theory,  where the morphisms $f$ and $g$ can represent arbitrary morphisms, in which case the {\em universal element} is the product element $a \otimes b$. 

Analogous to the universal construction of coproducts above, let us define the universal construction of a product in a category to show how it can be used to define more general types of universal graphical models. The commutative diagram of a limit is shown below, which can be viewed as a form of ``universal collider": 

\begin{center}
\begin{tikzcd}
  T
  \arrow[drr, bend left, "x"]
  \arrow[ddr, bend right, "y"]
  \arrow[dr, dotted, "r" description] & & \\
    & X  \times Y \arrow[r, "p"] \arrow[d, "q"]
      & X \arrow[d, "f"] \\
& Y \arrow[r, "g"] &Z
\end{tikzcd}
\end{center} 

This diagram asserts that there is a ``pullback" object labeled $X \times Y$ with morphisms $p: X \times Y \rightarrow X$ and $q: X \times Y \rightarrow Y$, which can be viewed as a generalization of the canonical projections from a cartesian product to its components. Furthermore, the diagram asserts that given any morphism from an object $T$ to $X$, there is a unique way to factor that morphism through the product object, so that the diagram ``commutes", meaning the morphism $x = p \circ r$. Similarly, any morphism from $T$ to $Y$ is also uniquely factored through $r$, so that $y = q \circ r$. We have thus characterized the product object purely in terms of the morphisms into and out of the object.  The use of universal constructions to define ``universal causal models" is explored in greater detail in \citep{sm:uc}. These constructions show how categoroids generalize graphoids. $\bullet$

\subsubsection{Integer Valued Multi-sets as Categoroids} 

Now, we turn to integer-valued multisets, or imsets, as described in \citep{studeny2010probabilistic}. Our goal is to explore the connections between imsets and categoroids. As mentioned earlier, imsets are defined as an integer-valued multiset function $u: \mathbb{Z}^{{\cal P(\mathbb{Z})}} \rightarrow \mathbb{Z}$ from the power set of integers, ${\cal P(\mathbb{Z})}$ to integers $\mathbb{Z}$. An imset is defined over partialy ordered set (poset), defined as a distributive lattice of disjoint (or non-disjoint) subsets of variables. The bottom element is denoted $\emptyset$, and top element represents the complete set of variables $N$. A full discussion of the probabilistic representations induced by imsets is given \citep{studeny2010probabilistic}. We will only focus on the aspects of imsets that relate to its conditional independence structure, and its topological structure as defined by the poset.  A {\em combinatorial} imset is defined as: 

\[ u = \sum_{A \subset N} c_A \delta_A \]

where $c_A$ is an integer, $\delta_A$ is the characteristic function for subset $A$, and $A$ potentially ranges over all subsets of $N$. An {\em elementary} imset is defined over $(a,b \CI A)$, where $a,b$ are singletons, and $A \subset N \setminus \{a, b\}$. A {\em structural} imset is defined as one where the coefficients can be rational numbers. For a general DAG model $G = (V, E)$, an imset in standard form \citep{studeny2010probabilistic} is defined as 

\[ u_G = \delta_V - \delta_\emptyset + \sum_{i \in V} (\delta_{\mbox{{\bf  Pa}}_i} - \delta_{i \cup \mbox{{\bf Pa}}_i}) \] 

\begin{figure}[h] 
 \caption{An illustration of an integer-valued multiset (imset) consisting of a lattice of subsets over three elements for representing conditional independences in DAG models. We will redefine imsets as convolutions of the $\zeta$ function with the m\"obius function $\mu$, which will clarify its relation to categoroids. \label{imset}}
\centering
\begin{minipage}{0.5\textwidth}
\includegraphics[scale=0.3]{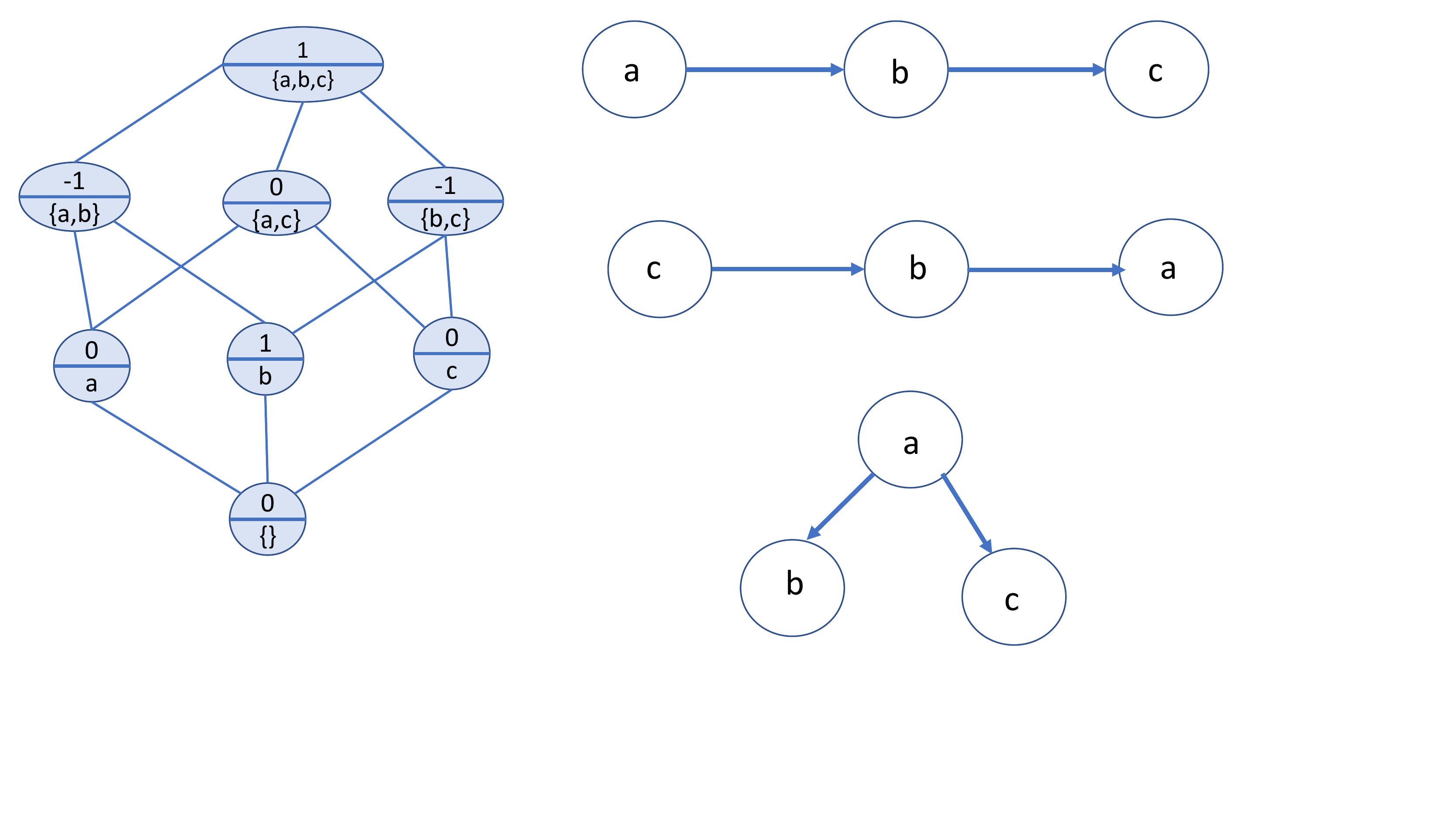}
\end{minipage} 
 \end{figure}

Figure~\ref{imset} shows an example imset for DAG models over three variables, defined by an integer valued function over the lattice of subsets. Each of the three DAG models shown defines exactly the same imset function.  \citet{studeny2010probabilistic} gives a detailed analysis of imsets as a non-graphical representation of conditional independence. We construct a novel representation of imsets using the theory of M\"obius inversion in this section, which we will then use to prove a theorem, showing that imsets define a special type of categoroid. 

\begin{definition}
The {\bf incidence algebra} over a poset $({\cal P}, \leq)$ is defined as the collection of all functions that are $0$ for all pairs of incomparable elements on the poset: 

\[ \Lambda({\cal P}) = \{f: {\cal P}^2 \rightarrow \mathbb{R} | f(x,y) = 0 \ \ \mbox{whenever} \ \ x \nleq y \} \] 
\end{definition}

We will show that imsets, which are essentially functions on posets, can be decomposed using a convolution of more elementary incidence algebra functions on posets. This reformulation, which can be viewed as a generalized Fourier ``change of basis" of imsets, uses the properties of M\"obius inversions. 

\begin{definition}
The {\bf zeta} function on posets is defined as $\zeta(x,y) = 1$ whenever $x \leq y$, and $0$ when $x \nleq y$. 
\end{definition}

\begin{definition}
The {\bf convolution} $f \star g$ of two functions on a poset $({\cal P}, \leq )$ is defined as: 

\[ f \star g (x,y) = \sum_{z: x \leq z \leq y} f(x,z) g(z,y)  \] 
\end{definition}

As an example, it is easy to show that for any function $f$ convolved with the $\zeta$ function produces a new function that is defined over the poset ideal of all elements $x \leq y$. This relationship will be central to the construction below. 

\begin{eqnarray*}
(f \star \zeta) (x,y) &=& \sum_{z: x \leq z \leq y} f(x,z) \zeta(z,y) \\
&=& \sum_{z: x \leq z < y} f(x,z) 
\end{eqnarray*}
 
\begin{definition}
 The {\bf M\"obius} function $\mu(x,y)$ is the convolutional inverse of the $\zeta(x,y)$ function: 
 
 \[ (\zeta(x,y) \star \mu(x,y)) = \delta(x,y) \] 
 
 where $\delta(x,y) = 1$ if and only if $x = y$. It is both a left and a right inverse. 
 \end{definition}
 
 The proof of the following result is standard in any combinatorics text, and will not be given. 
 
 \begin{theorem}
 \label{mobius}
 Let $({\cal P}, \leq)$ be any poset, and let $e: {\cal P} \rightarrow \mathbb{R}$. Suppose that ${\cal P}$ has a unique minimal element $\perp$. Define the following function $n: {\cal P} \rightarrow \mathbb{R}$ as follows, for any $a \in {\cal P}$: 
 
 \[ n(a) = \sum_{x \leq a} e(x) \] 
 
 We can then use the m\"obius function to ``invert" the original function, and perform a change of basis: 
 
 \[ e(a) = \sum_{x \leq a} n(x) \mu(x,a) \] 
 
 If we define $f(\perp,a) = e(a)$, and $f(x,y) = 0$ for all other undefined values, and $g(\perp,a) = n(a)$, where similarly $g(x,y) = 0$ for all other undefined values, then it follows that: 
 
 \[ g(\perp, a) = (f \star \zeta) (\perp, a) \] 
 
 \[ e(a) = (g \star \mu) (\perp, a) \] 
 \end{theorem}
 
 The importance of this theorem is that it shows we can reformulate any integer-valued multiset function using the m\"obius transform to a function defined over the incidence algebra of a poset. This change of basis will clarify the connection to the arrow structure of a categoroid. To the best of our knowledge, the following simple theorem below is novel in the literature. 
 
 \begin{theorem}
 Any integer-valued multiset (imset) function $u$ over a poset $({\cal P}, \leq)$ can be written in terms of a convolution of the $\mu$ function on the poset ${\cal P}$. 
 \end{theorem}
 
 {\bf Proof:} As shown by \cite{studeny2010probabilistic}, any integer-valued multiset function $u$ can be written as the linear combination of a set of simpler multiset functions, involving elementary, combinatorial, or structural multiset functions, with the only change being the coefficient structure. As this does not affect our proof, we simply assume that the function $u$ is defined as a linear combination of elementary multiset functions, such as: 
 
 \[ u = \sum_{A \subset N} c_A \delta_A \]
 
 Note that as $\delta_A$, the characteristic function for a set $A \subset N$, is a function on the poset ${\cal P}$, which is assumed to have a unique bottom element $\perp$, we can write it using Theorem~\ref{mobius} as: 
 
  \[ u(A) = \sum_{B \subset A} n(B) \mu(B,A) \] 
 
 where the auxiliary function $n$ is defined as: 
 
 \[ n(A) = \sum_{B \subset A} u(B) \] 
 
 Note that for the case of DAG models, such as the ones shown in Figure~\ref{imset}, the bottom element is the $\emptyset$, which is the unique bottom element in the lattice of subsets defining the graphoid axioms using which the theory of imsets is defined. In effect, this construction shows that any imset can be written as the pair of convolutions involving the $\mu$ mobius function and the $\zeta$ function: 
 
 \[ g(\perp, A) = (f \star \zeta) (\perp, A) \] 
 
 \[ u(A) = (g \star \mu) (\perp, A) \] 

The significance of this change of basis of the imset function is that it shows that each imset function can be written as a linear combinations of functions defined as arrows in a categoroid. This statement is the basis of the next theorem. $\bullet$

\begin{theorem}
Any integer-valued multiset (imset) function on a poset $({\cal P}, \leq)$ defines a categoroid ${\cal C}$. 
\end{theorem}

{\bf Proof:} As in the previous two cases of graphoids and separoids, the proof is quite straightforward, and involves showing how each of the elements in a categoroid can be defined using an imset: 

\begin{itemize} 

\item The set $O$ of objects in the categoroid ${\cal C}$ is the elements of the poset ${\cal P}$.  

\item The set $A$ of arrows is defined as the morphisms $f: x \rightarrow y$ if $x \leq y$ in the poset. In particular, for posets defined over a lattice of subsets, as shown in Figure~\ref{imset}, the partial order is defined by the subset relation on the lattice. 

\item The set of trigonoidal morphisms $T$ is defined using the conditional inference rules that are implicit in the definition of imsets. Following the construction above, each imset $u(A)$, where $A \subset N$ defines a function on the poset ideal $B \subset A$. These representation of each imset can be then transformed into a linear combination of morphisms $f$, where each morphism $f$ is defined using the m\"obius inversion of an imset. To capture inference rules involving conditional independence properties, we can, as before, include the axioms of (semi)graphoids as trigonoidal morphisms. To make this point clear, Lemma 6.1 in \cite{studeny2010probabilistic} states that if $u$ and $v$ are structural imsets, then a conditional independence inference implication $u \rightarrow v$ follows if and only if $\exists l \in \mathbb{N}$, such that l . u  - v is a structural imset. Each such property can be encoded as a trigonoidal morphism in the categoroid. 

\item The set of bridge arrows $B^0$ act as a connection between the poset relation $x \leq y$ to the conditional independence statement $x \CI x | y$.  Stated more generally, each bridge arrow here is written as $f: (x, y) \rightarrow (x, y, x)$ precisely when $f: x \rightarrow y$. 

\item Finally, the set of bridge arrows $B^1$ act as a connection between the conditional independence statement $x \CI x | y$ to the partial ordering $x \leq y$. Stated in more general terms, $g: (x, y, x) \rightarrow (x,y)$. These can remain the same as was done for separoids and graphoids. $\bullet$ 

\end{itemize}

\section{Functoroids and Natural Transformations over Categoroids} 

Our  goal is to construct a universal representation of any conditional independence, for which we define functoroids and natural transformations over functoroids. As we show later, these notions play a vital role in defining universal representations of conditional independences. We use a modified form of the Yoneda Lemma \citep{maclane:71}, which constructs a set-valued functor defined by the collection of all arrows (of all types) from an object, and uses a natural transformation from this set-valued functor to any other set-valued functor defined on a categoroid as a universal representation of that functor. In effect, we are constructing through the Yoneda Lemma a ``universal simulator" of a conditional independence structure, such as a graphoid or separoid, or possibly even a non-finite parameterization, such as proposed by \citet{studeny2010probabilistic}. Every proof of a conditional independence property that is deduced in a separoid by invoking a sequence of rules from {\bf P1} through {\bf P6} above is simulated in the categoroid by using the trigonoidal morphisms. 

\begin{definition}
A {\bf functoroid} $F: {\cal C} \rightarrow {\cal D}$ is a mapping from one categoroid to another, which comprises of the following mappings: 
\begin{itemize} 
\item For each object $o \in {\cal C}$, $F(o)$ is the corresponding object in ${\cal D}$. 

\item For each arrow $f: c \rightarrow d$ in ${\cal C}$, a corresponding arrow $F(f): F(c) \rightarrow F(d)$ in ${\cal D}$

\item For each trigonoidal arrow $f:(a, b, c) \rightarrow (a', b', c')$ in ${\cal C}$, a corresponding trigonoidal arrow $F(f):(F(a), F(b), F(c)) \rightarrow (F(a'), F(b'), F(c'))$ in ${\cal D}$. 

\item For each bridge arrow  $f:(x, y) \rightarrow (x, y, x)$, a corresponding bridge arrow $F(f):(F(x), F(y)) \rightarrow (F(x), F(y), F(x))$. 

\item Finally, for each bridge arrow $g:(x, y, x) \rightarrow (x,y)$, a corresponding bridge arrow $F(g):(F(x), Fy), F(x)) \rightarrow (F(x),F(y))$. 
\end{itemize} 
\end{definition}

It should be clear that functoroids can be composed easily, so that $G \circ F$ can be defined as the composite functoroid from ${\cal C}$ to ${\cal E}$, by composing the functoroids $F: {\cal C} \rightarrow {\cal D}$ and the functoroid $G: {\cal D} \rightarrow {\cal E}$, with the only caveat that bridge arrows are composed as defined earlier, in keeping with their asymmetric structures. 

\begin{definition}
The {\bf categoroid natural transformation} between two functoroids $F, G: {\cal C} \rightarrow {\cal D}$ is defined component-wise as follows: for each object $a$ in ${\cal C}$, and each morphism $f: a \rightarrow b$ in ${\cal C}$, the following diagram commutes:  

\begin{center}
\begin{tikzcd}
F(a) \arrow[r, "\eta_a"] \arrow[d, "f", red]
& D(a) \arrow[d, "f" red] \\
F(b) \arrow[r, red, "\eta_b" blue]
& |[blue]| D(b) 
\end{tikzcd}
\end{center} 

To capture the action of the trigonoids, we define a corresponding natural transformation for each trigonoid $f$ as: 

\begin{center}
\begin{tikzcd}
(F(a), F(b), F(c)) \arrow[r, "\lambda_{abc}"] \arrow[d, "f", red]
& (G(a), G(b), G(c)) \arrow[d, "f" red] \\
(F(a'), F(b'), F(c')) \arrow[r, red, "\lambda_{a'b'c'}" blue]
& |[blue]| (D(a'), D(b'), D(c'))
\end{tikzcd}
\end{center} 

Finally, to capture the action of the bridge arrows, we define a corresponding natural transformation for each $f \in B^0$ as: 

\begin{center}
\begin{tikzcd}
(F(a), F(b), F(c)) \arrow[r, "\lambda^0_{abc}"] \arrow[d, "f", red]
& (G(a), G(b), G(c)) \arrow[d, "f" red] \\
(F(a'), F(b')) \arrow[r, red, "\lambda^0_{a'b'}" blue]
& |[blue]| (D(a'), D(b'))
\end{tikzcd}
\end{center} 

and likewise, for each bridge arrow $f \in B^1$, we get: 
\begin{center}
\begin{tikzcd}
(F(a), F(b)) \arrow[r, "\lambda^1_{abc}"] \arrow[d, "f", red]
& (G(a), G(b)) \arrow[d, "f" red] \\
(F(a'), F(b'), F(c')) \arrow[r, red, "\lambda^1_{a'b'c'}" blue]
& |[blue]| (D(a'), D(b'), D(c'))
\end{tikzcd}
\end{center}

\end{definition}

Natural transformations compose as well between functoroids $F$, $G$ and $H$ all mapping categoroid ${\cal C}$ to ${\cal D}$. We can indicate the composed natural transformation abstractly as follows (where $\alpha$ and $\beta$ bundle together all the components of the natural transformation across normal, trigonoidal, and bridge arrows): 

\begin{center}
\begin{tikzcd}[row sep=huge]
    \mathcal{C}
     \arrow[r, bend left=65, "F"{name=F}]
     \arrow[r, "G"{inner sep=0,fill=white,anchor=center,name=G}]
     \arrow[r, bend right=65, "H"{name=H, swap}]
     \arrow[from=F.south-|G,to=G,Rightarrow,shorten=2pt,"\alpha"] 
     \arrow[from=G,to=H.north-|G,Rightarrow,shorten=2pt,"\beta"] &
   \mathcal{D}.
\end{tikzcd} 
\end{center} 

. 
\section{Further Examples of Categoroids} 

Here are a few of the many examples of categoroid structures, which have proposed in the literature. Many of these are discussed in \citep{DBLP:journals/amai/Dawid01}. \citep{witsenhausen:1975} proposed {\em information fields}, a measure-theoretic model of decision making that uses a lattice structure of sigma algebras to model decentralized decision-making in multi-agent systems. \citep{cif} apply Witsenhausen's model to define a topological version of conditional independence. \citep{sm:homotopy} defines conditional independence using Alexandroff finite topological spaces, and introduces the idea of constructing homotopic equivalences among causal models.

\begin{itemize} 

\item {\bf Modulo arithmetic:} The categoroid $(\mathbb{Z}, \equiv)$ defines modulo arithmetic, where the ternary relation is defined as $a \equiv b \ \mbox{mod} \ n$, for $a, b, n \in \mathbb{Z}$.  The integers $\mathbb{Z}$ form a semi-lattice, where the join $\vee$ represents least upper bound between two integers. The trigonoids are defined by the $\equiv$ congruence relation between triples of numbers, so for example, $a \equiv b  \ \mbox{mod} \ n \Rightarrow b \equiv a  \ \mbox{mod} \ n $. It is interesting to note that the m\"obius inversion procedure applies in this case to give an elegant characterization of the prime factorization property of integers.

\item {\bf Finite Space Topological Categoroids:} A categoroid can be defined over finite space topologies, where the trigonoids are defined as $(B \CI_t C | A)$ denoting  {\em topological conditional independence}, as discussed in \citep{cif,sm:homotopy}. The preorder relation can be defined over any (finite or not) topology in terms of containment over open sets.   Topological conditional independence is  defined using the notion of continuity of paths over finite (Alexandroff) topological spaces, where a path $p$ is continuous if it can be represented by a continuous function from the unit interval $(0, 1)$ to $p$. We give the details of  this construction for finite Alexandroff spaces in the next section. 

\item {\bf Orthogonoids in Hilbert and Inner Product Spaces:} \citep{DBLP:journals/amai/Dawid01} defines orthogonoids in inner product and Hilbert spaces. A subspace $X$ is {\em orthogonal independent of} $Y$ given $Z$, denoted $X \CI_O Y | Z$ if in some inner product space $(I, \langle . \ , \ . \rangle)$, the projection $\Pi_{X \vee Z} x$ onto $Y \vee Z$ is lies in $Z$. 

\item {\bf Sigma Fields and Probability Spaces:} Let $(\Sigma, {\cal F}, P)$ denote a probability space, and let ${\cal S}$ denote the lattice of sub-$\sigma$ fields of ${\cal F}$, ordered by inclusion. For ${\cal A}, {\cal B}, {\cal C} \in {\cal S}$, \citet{DBLP:journals/amai/Dawid01} defines ${\cal A} \CI_p {\cal B} | {\cal C} [P]$ to denote that sigma fields ${\cal A}$ and ${\cal B}$ are (conditionally) independent, given ${\cal C}$ (under the probability measure $P$). If ${\cal C} = \{ \emptyset, \Omega \}$ is the trivial $\sigma$-field, then conditional independence becomes the usual property that ${\cal A}$ is marginally independent of ${\cal B}$ (under $P$). This notion of conditional independence can be seen as a basis of the work on causal information fields \citep{cif}, which is based on Witsenhausen's definition of information fields \citep{witsenhausen:1975}, which are subfields of a product sigma algebra over a collection of decision variables. 

\item {\bf Graphoids:} \citet{pearl:bnets-book} introduced {\em graphoids}, which can be viewed as a strong separoid over a distributive lattice, which contains a minimal $0$ element, and posesses relative complements, and restricts the join $\vee$ operation to pairwise disjoint terms. 

\item {\bf Graphical models:} Various type of graphical models, from DAGs \citep{pearl:bnets-book} to marginalized DAGs \citep{mdag}, chain graphs \citep{lauritzen:chain},  hyperedge-directed graphs \citep{hedge}, and lattice conditional independence models \citep{DBLP:conf/uai/AnderssonMP96} can all be viewed as categoroids, where the conditional independence structure is encoded in the structure of the graph. Later we will formally define the relationship between graphical models and categoroids using adjunctions on categoroids. 

\item {\bf Co-Presheaves:} Co-presheaves define the functoroid category $\hat{C} =$ {\bf Set}$^{\cal C}$. The Yoneda embedding of any categoroid ${\cal C}$ into the categoroid  {\bf Set} yields a presheaf, where each object $c$ in the category is mapped to the functor {\bf Hom}$_{\cal C}(c,-)$ of the set of all (regular, bridge, and trigonoidal) morphisms into object $c$. We will derive a novel form of the Yoneda Lemma for categoroids below, building on the definition of the categoroid natural transformation above. 

\item {\bf Commutative monoidal preorders over co-presheaves:} A {\em commutative monoidal preorder} is a preorder $({\cal S}, \leq)$ and a commutative monoid $({\cal C}, \otimes, {\bf 1})$ satisfying the property $x \otimes y \leq x' \otimes y'$ whenever $x \leq x'$ and $y \leq y'$. \citep{Bradley_2022} studied commutative monoidal preorders over co-presheaves applied to deep language models, where objects represents partial sentences in English and morphisms represent all possible completions. 

\item {\bf Gr\"obner categories:} \citet{Sam_2016} introduced {\em Gr\"obner} categories. In brief, in a Gr\"obner category, for each object $x \in C$, a representation $P_x$ is defined by $P_x(y) = {\bf k}[{\bf Hom}_{\cal C}(x,y)]$ as the free {\bf k} module with basis {\bf Hom}$_{\cal C}(x,y)$. The set of isomorphism classes of representations $|{\cal C}_x|$ must admit an admissible order, and the poset representation induced by ${\cal C}_X$ must be noetherian, that is, satisfy the ascending chain condition. 
\end{itemize}

\section{Universal Constructions in Categoroids} 

The principal aim of this paper is to elucidate {\em universal} properties of conditional independence in categoroids. We need to precisely define what is meant by the term ``universal". We follow the definition given by \citet{riehl2017category} below. We first need to define some terminology from category theory. 

\subsection{Covariant and Contravariant Functoroids} 

Our goal is to abstract from the previous representations of conditional independence to define universal representations of categoroids using a novel variant of the Yoneda Lemma.   To do that, we need to introduce some preliminary background material for readers unfamiliar with some basic terminology. Category theory can be viewed as the ``science of analogy". Instead of asking the question whether two objects are ``equal", it instead poses the question of whether objects are {\em isomorphic}. The Yoneda Lemma shows how to construct universal representations of objects in a category, so that they are fully and faithfully embedded up to isomorphism in the category of sets. 

\begin{definition}
Two objects $X$ and $Y$ in a categoroid ${\cal C}$ are deemed {\bf isomorphic}, or $X \cong Y$ if and only if there is an invertible morphism $f: X \rightarrow Y$, namely $f$ is both {\em left invertible} using a morphism $g: Y \rightarrow X$ so that $g \circ f = $ {\bf id}$_X$, and $f$ is {\em right invertible} using a morphism $h$ where $f \circ h = $ {\bf id}$_Y$. 
\end{definition}

Functoroids come in two varieties. 

\begin{definition} 
A {\bf covariant functoroid} $F: {\cal C} \rightarrow {\cal D}$ from categoroid ${\cal C}$ to categoroid ${\cal D}$, and defined as the following: 
\begin{itemize} 
    \item An object $F X$ (sometimes written as $F(x)$ of the categoroid ${\cal D}$ for each object $X$ in categoroid ${\cal C}$.
    \item An regular arrow  $F(f): F X \rightarrow F Y$ in categoroid ${\cal D}$ for every arrow  $f: X \rightarrow Y$ in categoroid ${\cal C}$. 
    \item A trigonoidial arrow $F(f): F X \rightarrow F Y$ in categoroid ${\cal D}$ for every trigonoidal arrow  $f: X \rightarrow Y$ in categoroid ${\cal C}$, where $X$ and $Y$ now denote trigonoidal objects (triples). 
     \item A bridge arrow $F(f): F X \rightarrow F Y$ in categoroid ${\cal D}$ for every bridge arrow  $f: X \rightarrow Y$ in categoroid ${\cal C}$, respecting the type of bridge morphism (for example, if $f \in B^0$, then $X \in O \times O$ and $Y \in O \times O \times O$, and if $f \in B^1$, then $X \in O \times O \times O$, and $Y \in O \times O$). 
   \item The preservation of identity and composition: $F \ id_X = id_{F X}$ and $(F f) (F g) = F(g \circ f)$ for any composable arrows $f: X \rightarrow Y, g: Y \rightarrow Z$ of any type (keeping in mind that bridge arrows $B^0$ compose only with $B^1$ arrows, and vice versa). 
\end{itemize}
\end{definition} 

\begin{definition} 
A {\bf contravariant functoroid} $F: {\cal C} \rightarrow {\cal D}$ from category ${\cal C}$ to category ${\cal D}$ is defined exactly like the covariant functoroid, except all the arrows are reversed. In the contravariant functoroid $F: C^{\mbox{op}} \rightarrow D$, every morphism $f: X \rightarrow Y$ is assigned the reverse morphism $F f: F Y \rightarrow F X$ in category ${\cal D}$. Similarly, the trigonoidal arrows are also reversed. Particularly important is the way contravariance works for bridge arrows. The contravariant version of the $B^0$ arrow turns it into a $B^1$ arrow, and similarly, the contravariant version of the $B^1$ arrow turns it into a $B^0$ arrow (this follows from the asymmetry in their domain and range). 
\end{definition} 

We introduce the following functors that will prove of value in the proof of the Yoneda Lemma: 

\begin{itemize} 
\item For every object $X$ in a categoroid ${\cal C}$, there exists a covariant functoroid ${\cal C}(X, -): {\cal C} \rightarrow {\bf Set}$ that assigns to each object $Z$ in ${\cal C}$ the set of morphisms ${\cal C}(X,Z)$, and to each morphism $f: Y \rightarrow Z$, the pushforward mapping $f_*:{\cal C}(X,Y) \rightarrow {\cal C}(X, Z)$. Similarly, for every trigonoidal object $(x,y,z)$ in ${\cal C}$, there is a covariant functoroid ${\cal C}((x,y,z),-): {\cal C} \rightarrow {\bf Set}$ that assigns to each object $(x',y',z')$ in ${\cal C}$, the trigonoidal morphisms ${\cal C}((x,y,z),(x',y',z'))$, and to each trigonoidal morphism $f: (x",y",z") \rightarrow (x',y',z')$ the pushforward $f_*: {\cal C}((x,y,z),(x",y",z")) \rightarrow {\cal C}((x,y,z),(x',y',z'))$. Similarly, for every object $(x,y,z)$ in ${\cal C}$, there is a covariant functoroid ${\cal C}((x,y,z),-): {\cal C} \rightarrow {\bf Set}$ that assigns the bridge morphisms ${\cal C}((x,y,z),(x',y'))$, and to each bridge morphism $f: (x",y") \rightarrow (x',y',z')$ the pushforward $f_*: {\cal C}((x,y,z),(x",y")) \rightarrow {\cal C}((x,y,z),(x',y',z'))$. 

\item For every object $X$ in a category ${\cal C}$, there exists a contravariant functoroid ${\cal C}(-, X): {\cal C}^{\mbox{op}} \rightarrow {\bf Set}$ that assigns to each object $Z$ in ${\cal C}$ the set of morphisms {\bf Hom}$_{\cal C}(X,Z)$, and to each morphism $f: Y \rightarrow Z$, the pullback mapping $f^*:$ {\bf Hom}$_{\cal C}(Z, X) \rightarrow {\cal C}(Y, X)$. Note how ``contravariance" implies the morphisms in the original category are reversed through the functorial mapping, whereas in covariance, the morphisms are not flipped. Similarly, we need to define contravariant trigonoidal morphisms and bridge morphisms, following the above constructions, with the only proviso again noting that contravariant bridge morphisms of one type get converted into the other type, as noted earlier. 
\end{itemize} 

\begin{definition} 
\label{fully-faithful} 
Let ${\cal F}: {\cal C} \rightarrow {\cal D}$ be a functoroid from categoroid ${\cal C}$ to categoroid ${\cal D}$. If for all arrows $f$, including regular, trigonoidal and bridge, the mapping $f \rightarrow F f$
\begin{itemize}
    \item injective, then the functoroid ${\cal F}$ is defined to be {\bf faithful}. 
    \item surjective, then the functoroid ${\cal F}$ is defined to be {\bf full}.  
    \item bijective, then the functoroid ${\cal F}$ is defined to be {\bf fully faithful}. 
\end{itemize}

\end{definition}

\subsection{Generalizing Joins to Co-limits} 

As illustrated earlier, we can generalize joins to co-products and co-limits, providing a way to generalize the notion of conditional independence based on join structures in separoids. As we defined co-products above, we turn to define colimits. 

\begin{definition}
Given a functoroid $F: {\cal J} \rightarrow {\cal C}$ from an indexing diagram category ${\cal J}$ to a categoroid ${\cal C}$, an element $A$ from the set of natural transformations $N(A,F)$ is called a {\bf cone}. A {\bf limit} of the diagram $F: {\cal J} \rightarrow {\cal C}$ is a cone $\eta$ from an object lim $F$ to the diagram satisfying the universal property that for any other cone $\gamma$ from an object $B$ to the diagram, there is a unique morphism $h: B \rightarrow \mbox{lim} F$ so that $\gamma \bullet = \eta \bullet h$ for all objects $\bullet$ in ${\cal J}$. Dually, the {\bf co-limit} of the diagram $F: {\cal J} \rightarrow {\cal C}$ is a cone $\epsilon$ satisfying the universal property that for any other cone $\gamma$ from the diagram to the object $B$, there is a unique mapping $h: \mbox{colim} F \rightarrow B$ so that $\gamma \bullet = h \epsilon \bullet$ for all objects $\bullet$ in ${\cal J}$. 
\end{definition}

\subsection{Kan Extensions over Categoroids} 

Kan extensions are the single most powerful universal construction in category theory from which every other concept can be defined. \citet{maclane:71} stated it boldly as ``Every concept is a Kan extension". It is well known in category theory that ultimately every concept, from products and co-products, limits and co-limits, and ultimately even the Yoneda embeddings, can be derived as special cases of the Kan extension \citep{maclane:71}. Kan extensions are usually defined over categories, but since they are stated in terms of natural transformations, they can readily be generalized to categoroids.  Kan extensions intuitively are a way to approximate a functoroid ${\cal F}$ so that its domain can be extended from a categoroid ${\cal C}$ to another categoroid  ${\cal D}$.  Because it may be impossible to make commutativity work in general, Kan extensions rely on natural transformations to make the extension be the best possible approximation to ${\cal F}$ along ${\cal K}$. 

\begin{definition}
A {\bf left Kan extension} of a functoroid $F: {\cal C} \rightarrow {\cal E}$ along another functoroid $K: {\cal C} \rightarrow {\cal D}$, is a functoroid $\mbox{Lan}_K F: {\cal D} \rightarrow {\cal E}$ with a natural transformation $\eta: F \rightarrow \mbox{Lan}_F \circ K$ such that for any other such pair $(G: {\cal D} \rightarrow {\cal E}, \gamma: F \rightarrow G K)$, $\gamma$ factors uniquely through $\eta$. In other words, there is a unique natural transformation $\alpha: \mbox{Lan}_K F \implies G$. \\
%
%\arrow{rr, dotted}{\mbox{Lan}_K F}
\begin{center}
\begin{tikzcd}[row sep=2cm, column sep=2cm]
% drawing 0- and 1-celss
\mathcal{C}  \ar[dr, "K"', ""{name=K}]
            \ar[rr, "F", ""{name=F, below, near start, bend right}]&&
\mathcal{E}\\
& \mathcal{D}  \ar[ur, bend left, "\text{Lan}_KF", ""{name=Lan, below}]
                \ar[ur, bend right, "G"', ""{name=G}]
                
%
% drawing 2-cells  
\arrow[Rightarrow, "\exists!", from=Lan, to=G]
\arrow[Rightarrow, from=F, to=K, "\eta"]
\end{tikzcd}
\end{center}
\end{definition}

A {\bf right Kan extension} can be defined similarly.

\section{Yoneda Lemma for Categoroids}

In this section, we construct a universal representation of categoroids using a modified version of the well-known Yoneda Lemma \citep{maclane:71}.  The central philosophy underlying category theory is construct representations of objects in terms of their interactions with other objects. Unlike set theory, where an object like a set is defined by listing its elements, in category theory objects have no explicit internal structure, but rather are defined through the morphisms that define their interactions with respect to other objects. The celebrated Yoneda lemma makes this philosophical statement more precise. We state it first for categories, before proving an extended version for categoroids. 

\begin{theorem}
{\bf Yoneda Lemma:} For every object $X$ in category ${\cal C}$, and every covariant functor $F: {\cal C} \rightarrow {\bf Set}$, the set of natural transformations from ${\cal C}(X, -)$ to $F$ is isomorphic to $F X$. 
\end{theorem}

That is, the natural transformations from ${\cal C}(X, -)$ to $F$ serve to fully characterize the object $F X$ up to isomorphism. In the special circumstance when the set-valued functor $F = {\cal C}(Y, -)$, the Yoneda lemma asserts that $\mbox{Nat}({\cal C}(X, -), {\cal C}(Y, -) \cong {\cal C}(X, Y)$. In other words, a pair of objects are isomorphic $X \cong Y$ if and only if the corresponding contravariant functors are isomorphic, namely ${\cal C}(X, -) \cong {\cal C}(Y, -)$.

\begin{figure}[h]
\begin{center}
\begin{minipage}{0.5\textwidth}
\includegraphics[scale=0.4]{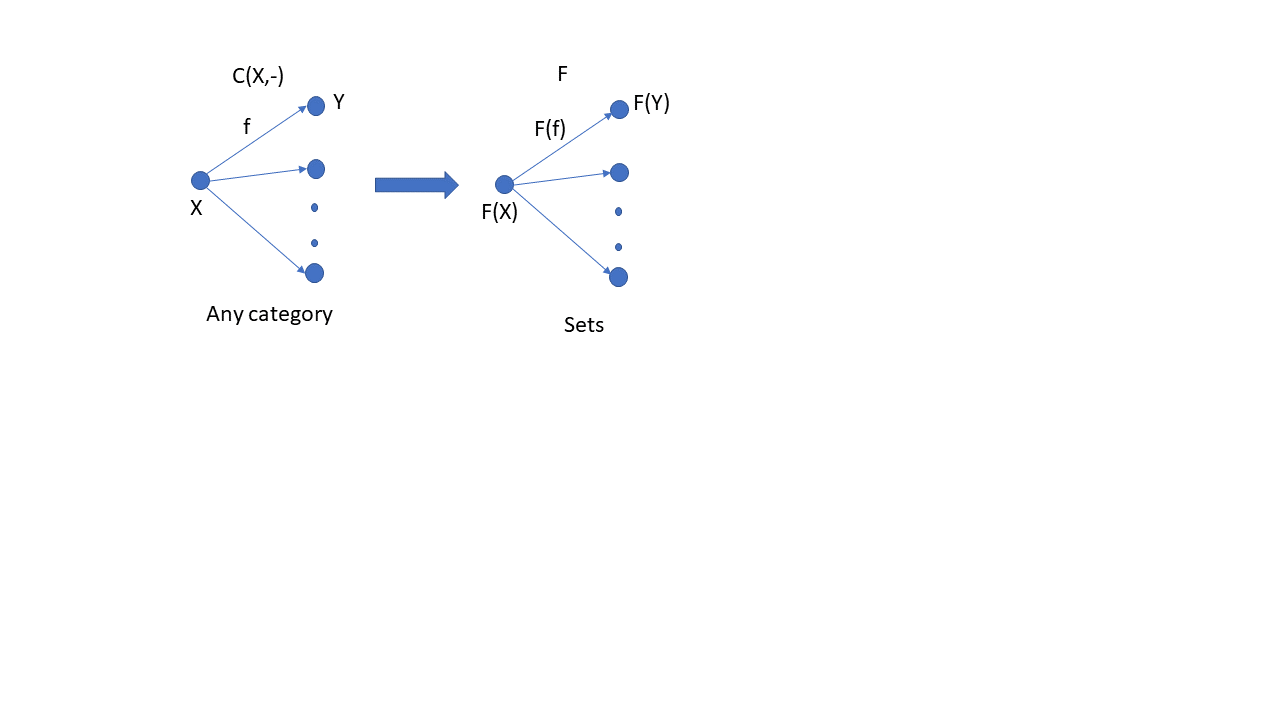}
\end{minipage} \vspace{-1.5in}
\end{center} 
\caption{The Yoneda Embedding shows how to construct a full and faithful representation of any set-valued functor $F$ into the category {\bf Sets}. The key idea is to recognize that $F$ must act functorially, and therefore it must map each object $X$ to a set $F(X)$, and each morphism $f: X \rightarrow Y$ into a function that maps the set $F(X)$ to $F(Y)$. To ``mimic" the action of the functor on $F$, it is sufficient to construct a ``basis" of elements in the category given by the set of all morphisms ${\cal C}(X, -)$ out of a category, which provides a representable functor. \label{yoneda-embedding}}
\end{figure}

To give some insight into the theorem, let us first understand at a high level how the proof of the original Yoneda Lemma works. The key insight in Yoneda Lemma is recognizing that any set-valued functor $F$ on a category must act {\em functorially}, that is, it must not only map any object $C \in {\cal C}$ to the set $F(C)$, but it must also map each morphism $f: C \rightarrow C'$ to the set-valued function $F(f): F(C) \rightarrow F(C')$ (see Figure~\ref{yoneda-embedding}). It seems almost impossible that no matter what the functor $F$ is, it is possible to ``mimic" the action of this functor using just the morphisms leaving an object $C$ as a universal representation. To get some intuition underlying the proof, note that for any element $x \in F(C)$, the function $F(f)$ must map $x$ into some element $y$ in $F(C')$. In other words, the action of $F(f)$ is simply the functional mapping of $x$ to $y$. Given that two sets of equal cardinality can be placed into bijective correspondence, it is sufficient to guarantee that there are enough elements in the morphism functor ${\cal C}(C, -)$ that can ``mimic" the action of the functor $F$. A bit of introspection reveals that this must be the case because the set-valued function $F(f)$ must map every element $x \in F(C)$, and there are exactly as many of those as there are morphisms coming out of $C$ in the category ${\cal C}$. 

\subsection{Generalizing the Yoneda Lemma to Categoroids} 

\begin{figure}[h]
\begin{center}
\begin{minipage}{0.5\textwidth}
\includegraphics[scale=0.4]{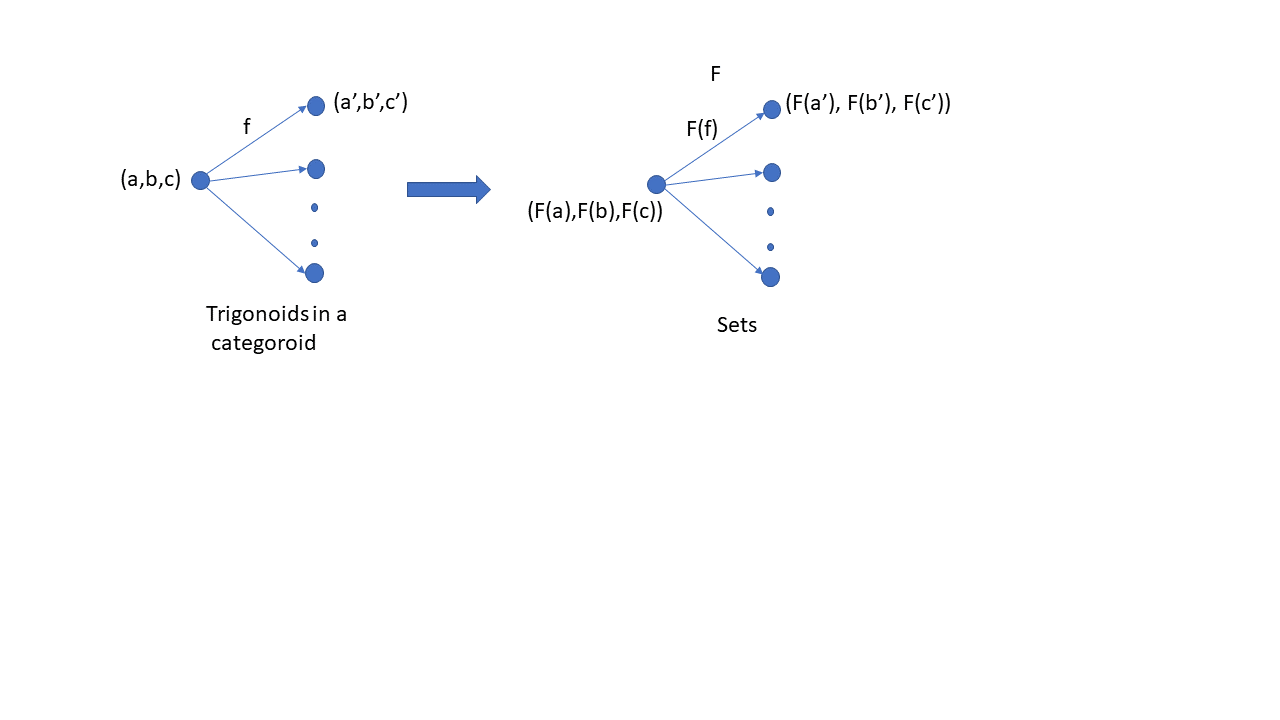}
\end{minipage} \vspace{-1.5in}
\end{center} 
\caption{The Yoneda Embedding for categoroids applies the same idea shown in Figure~\ref{yoneda-embedding} to construct a basis for representing trigonoid morphisms, where each morphism is now a mapping from one conditional independence triple $(a, b, c)$ into a new one $(a', b', c')$ that follows from applying any of the trigonoid morphisms in the categoroid. This process has to be repeated for bridge morphisms as well. \label{trigonoid-embedding}}
\end{figure}

The principal aim of this section is to show an enhanced variant of the Yoneda Lemma, which applies to categoroids. The principal difference, of course, is that categoroids are a join of two categories, one of which includes regular objects and morphisms, to which the above Yoneda Lemma directly applies, but the other includes trigonoids, morphisms that act over triples of objects, and there are  bridge morphisms defining the join. The modifications to the Yoneda Lemma are simple, because as we have defined the categoroid completely in terms of three types of morphisms, and have defined an augmented natural transformation over these three types of arrows, we just have to use the modified notion of natural transformation in defining the Yoneda Lemma for categoroids.  We give a detailed proof of the modified Yoneda Lemma, leaving aside some details that are covered in standard textbooks \citep{richter2020categories}.  Figure~\ref{trigonoid-embedding} gives the high level idea of constructing once again a basis for the trigonoid actions in the category of ${\bf Sets}$, which mimics the actions of the trigonoid morphisms in the original category. The details are given in the proof below. 

\begin{definition}
A functoroid $F: {\cal C} \rightarrow {\bf Sets}$ is {\bf representable} if it is isomorphic to a morphism functoroid, that is, there exists an object $C$ of {\cal C}, and a natural isomorphism over functoroids such that: 
\begin{equation}
    \eta_{C,F}: {\cal C}(C, -) \Rightarrow F
\end{equation}
\end{definition}

We now state the Yoneda Lemma for categoroids, and work through a detailed proof of it, mirroring the proof of the original Yoneda Lemma, noting places where it needs to be modified to account for the trigonoidal and bridge morphisms. Our notation below follows that in the book by \citet{richter2020categories}, which of course states the proof just for the regular case of categories, not categoroids. Much of the hard work of extending this lemma has already been accomplished above in our extended definitions of functoroids and categoroid natural transformations. 

\begin{theorem}
(Yoneda Lemma for Categoroids:) Let ${\cal C}$ be a categoroid and $F$ be a functoroid from the categoroid ${\cal C}$ to the category {\bf Sets}. 
\begin{itemize} 
\item For each object $C$ of ${\cal C}$, there is a bijection $Y(F,C)$ between the set of all categoroid natural transformations from ${\cal C}(C,-)$ to $F$, denoted as {\bf Nat}$({\cal C}(C, -), F)$, and the set $F(C)$. These include the regular objects $c \in {\cal C }$ as well as the trigonoidal objects $(a,b,c) \in {\cal C}$ defining conditional independence. 
\item The bijections $(Y_F)_C$ defined as $Y(F,C)$ are the components of a categoroid natural transformation $Y_F: {\bf Nat}({\cal C}(-,-), F) \Rightarrow F)$. 
\item If ${\cal C}$ is a small categoroid, that is, its collection of all (regular, bridge, trigonoidal) arrows are representable as a set, then the bijections $(Y_C)_F = Y(F,C)$ are the components of a categoroid natural transformation from the functoroid
\[ {\bf Nat}({\cal C}(C, -), -): \mbox{Fun}({\cal C}, {\bf Sets}) \rightarrow {\bf Sets} \] 
to the functoroid $\epsilon_C: \mbox{Fun}({\cal C}, {\bf Sets}) \rightarrow {\bf Sets}$ that sends each $F$ to the set $F(C)$. 
\end{itemize} 
\end{theorem}

{\bf Proof:} We will give a detailed proof of the Yoneda Lemma, including the extra components for dealing with categoroids, for the sake of completeness, and because the construction details will be of value later in the paper. 
Let us define the categoroid natural transformation $\eta_{C'}$ for an object $C'$ in {\cal C} by the function: 

\[ {\cal C}(C, C') \rightarrow F(C') \] 

That is, for each element $y \in F(C')$, we ``label" it with a morphism in ${\cal C}(C, C')$. Note there must be exactly as many (regular, bridge, trigonoidal) arrows in ${\cal C}(C, C')$ as there are elements in $F(C')$, by the definition of a functorial embedding $F$ into {\bf Sets}.  That is, if $F$ acts as a functoroid, it has to be map all the structure of a categoroid into {\bf Sets}, and thus, there must be for each regular arrow $f$, a corresponding function $F(f)$, for each trigonoidal arrow $t$, a trigonoidal function $F(t)$ over triples of elements in {\bf Sets}, and for each bridge arrow $b$ of the specific $B^0$ or $B^1$ type, a corresponding set-valued $F(b$ embedding in {\bf Set} (see Figure~\ref{trigonoid-embedding}).  

In other words, the functoroid $F$ must provide a means to carry out conditional independence inferences in the category ${\bf Set}$ to faithfully embed the original categoroid. As a specific example, if the functoroid $F$ is implemented by embedding a categoroid into a particular categoroid of graphical models of a particular class, e.g., undirected or directed graphs, then it must also enable all conditional independence inferences (which are usually  defined as set-valued operations on the graph structures). In effect, the Yoneda construction generate a ``basis" space that can mimic the effect of the functoroid $F$ on all three types of arrows in the category {\bf Sets}, exactly analogous to the Yoneda Lemma for regular categories. 

We now define $Y(F,C)$ as

\[ Y(F,C)(\eta) = \eta_C({\bf 1}_C) \] 

In other words, we are representing the action of the bijections using the action on the unit morphism on every object $C$. The reason for this will become clearer below, because we want to show that the Yoneda Embedding provides a full and faithful embedding (i.e., an isomorphism), so we want to be able to show that the composed mapping from the category into {\bf Sets} becomes a bijection. Essentially, what we have constructed is an evaluation functional that takes a morphism from $C$ to $C'$ in the category ${\cal C}$ and an element $x$ in the set embedding $F(C)$ to produce an element $f(x)$ in $F(C')$: 

\[ {\cal C}(C, C') \times F(C) \rightarrow F(C') \] 

For a fixed $x \in F(C)$, let us denote for any $f \in {\cal C}(C, C')$, this mapping as: 

\[ \tau(F, C)_{x, C'}(f) = F(f)(x) \] 

and repeating this for every possible $f$ gives us the following function: 

\[ \tau_{x, C'}(f) = F(f)(x) \] 

In other words, because we have chosen to ``code" each element $x \in F(C)$ by a suitable morphism in ${\cal C}(C, C')$, we can set up a 1-1 correspondence between them. This process ends up giving us a categoroid natural transformation

\[ \tau(F, C)_x: {\cal C}(F, -) \Rightarrow F \].

To show that the categoroid natural transformation we have constructed is composable, let us ``walk forward" an extra step and examine a morphism $g \in {\cal C}(C', C")$ that we can combine with the $f$ above, giving us: 

\[ F(g) (F(f)(x))  = F(g \circ f)(x)  = \tau(F, C)_{x, C"} \circ {\cal C}(C, g)(f) \] 

which essentially implies that 

\[ F(g) \circ \tau(F, C)_{x, C'} = \tau(F, C)_{x, C"} \circ {\cal C}(C, g) \] 

In summary, we can show the maps $Y(F, C)$ and $\tau(F,C)$ are inverses of each other for all types of arrows. For any $x \in F(C)$, we get: 

\[ Y(F,C)(\tau(F, C)_x) = \tau(F,C)_{x,C}({\bf 1}_C) = F({\bf 1}_C)(x) = {\bf 1}_{F(C)}(x) = x \] 

Conversely, for any categoroid natural transformation $\eta: {\cal C}(C, -) \Rightarrow F$, and for every $f: C \rightarrow C'$, it follows that: 

\[ \tau(F,C)_{Y(F,C)(\eta), C'}(f) = \tau(F,C)_{\eta_C({\bf 1}_C,C'}(f) = F(f)(\eta_C({\bf 1}_C) = \eta_{C'}(f) \] 

The same process can be repeated for showing the maps $Y(F,C)$ and $\tau(F,C)$ are inverses of each other for trigonoidal morphisms as well. Since this is just a tedious repetition using the conditional independence rules, we will skip this portion instead. 

There are two other parts to the proof of the Yoneda Lemma, one showing the naturality of the embedding over objects in ${\cal C}$, and the other showing that when ${\cal C}$ is a small category, the collection of morphisms is just a set, and the bijection can be realized in terms of set morphisms.  These steps are exactly analogous to the standard Yoneda Lemma in standard textbooks, e.g. \citep{richter2020categories}, and will not be repeated here. $\bullet$

\section{Universal Representations of Categoroids} 

One of the most significant themes in category theory is the concept of a {\em universal property}, which is discussed in Chapter 2 of the book by \citet{riehl2017category}. Our explanation follows her presentation. Universality at its core is also tied to the notion of {\em representability}. To say a property is universal is equivalent to saying it is representable through a so-called {\em universal element}.

\begin{figure}[h]
\centering
\vskip -0.35in
\begin{minipage}{0.5\textwidth}
\centering
\includegraphics[scale=0.35]{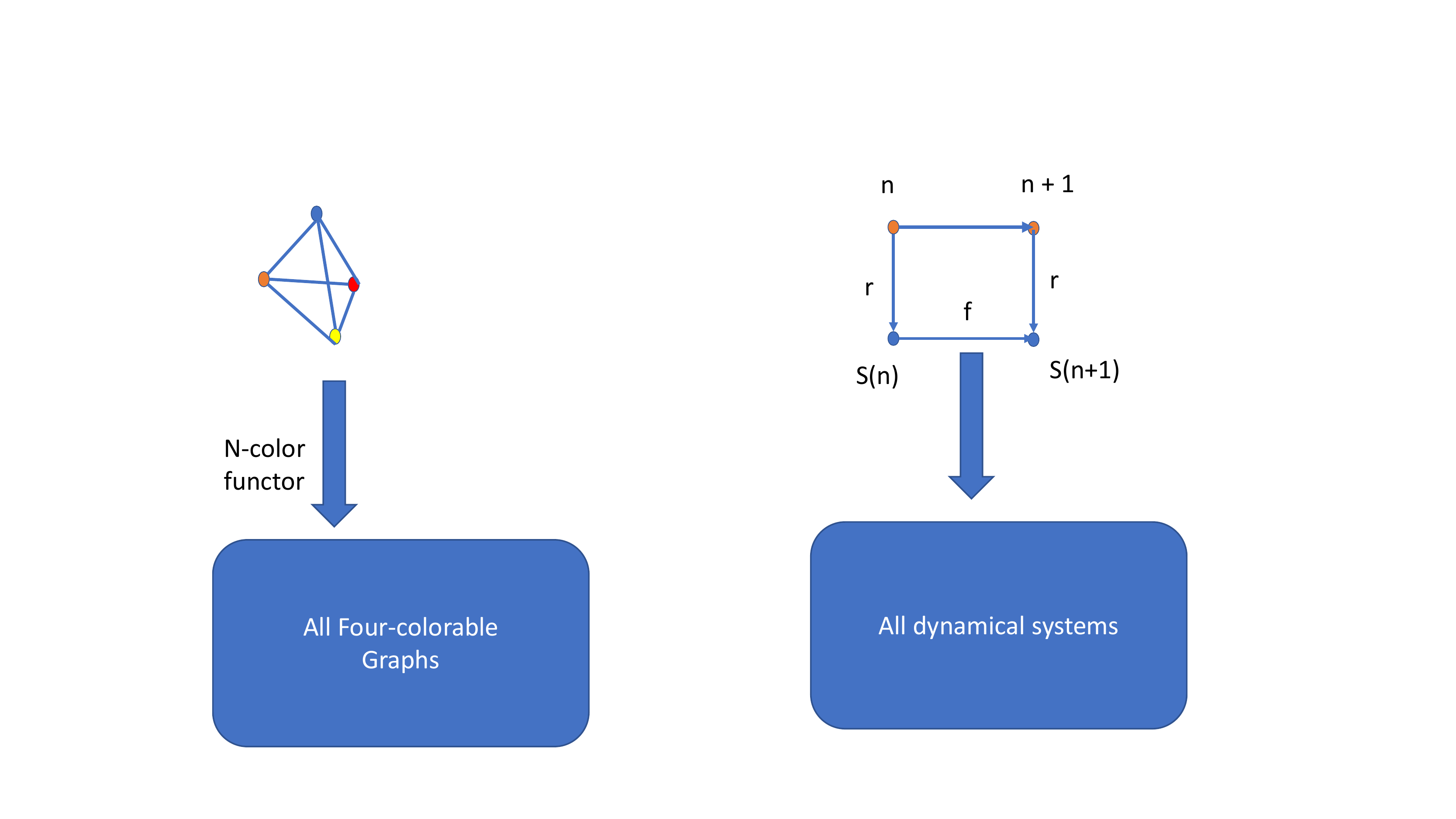}
\end{minipage}
\caption{To say a property is universal is to imply that it is functorially representable by a {\em universal element}. The $N$-color functor maps any undirected graph $G$ from the category of graphs ${\cal G}$ to the category {\bf Set} representing its coloring such that no two adjacent vertices have the same color. The complete graph $K_N$ on $N$ vertices is the  graph with the smallest number of edges and vertices that can be colored with no less than $N$ colors. Consequently, it represents the universal element for the functor $N$-color that maps any graph to the set of colors that can be used to color the graph. Similarly, the natural numbers $\mathbb{N}$ is the {\em universal dynamical system} because every other dynamical system can be expressed by a morphism from it.} 
\label{universal-element}
\end{figure}

 The concept is explained in Figure~\ref{universal-element}. Consider the functor $N$-color: {\bf Graph} $\rightarrow {\bf Set}$ that maps from  category of all undirected graphs to the category of sets, such that $N$-color(G) is defined as the set of colors that can color the vertices of the graph such that no two adjacent vertices get the same color. Clearly, $K_N$, the complete graph on $N$ vertices is the {\em smallest} graph in terms of number of edges and vertices that can be colored with no fewer than $N$ colors. For example, for $N=4$, the famous Four-color theorem tells us that all planar graphs of any size can be colored with 4 colors. So, the complete graph $K_4$ shown is the universal element that {\em represents} the universal property of $4$-colorability. Similarly, let us consider the category of all discrete dynamical systems defined by an endomorphism $f: S \rightarrow S$, where $S$ is a finite set, and $S_0$ is some distinguished initial object. The natural numbers $\mathbb{N}$ under the successor function $s: \mathbb{N} \rightarrow \mathbb{N}$ defined as $s(n) = n+1$ is the {\em universal discrete dynamical system}, because given any discrete dynamical system, we can define a unique morphism $r: \mathbb{N} \rightarrow S$ such that the diagram on the right in Figure~\ref{universal-element} commutes, namely $S(n) = r(n)$. The fundamental result from the Yoneda Lemma for categoroids show that the Yoneda embedding provides the universal representation for categoroids.

\subsection{The Density Theorem for Sheaves over Categoroids} 

The Yoneda Lemma provides a universal representations of categoroids by constructing representations in the functoroid category of (co)presheaves, where the universality is a consequence of deep properties of the sheaf structure itself. In other words, the universality is due entirely to the fact that it takes place in a presheaf, which is a topos \citep{maclane:sheaves}. 
The Yoneda Lemma defines the universal property of any object $X$ in a categoroid ${\cal C}$ in terms of a universal element that also serves to represent it, namely the copresheaf covariant functoroid {\bf Hom}$_{\cal C}(X, -)$ of all (types of) morphisms out of $X$. The copresheaf is an element of the functor category {\bf Set}$^{{\cal C}}$. 

One application of the Yoneda Lemma connects it to the use of abstract diagrams. 

\begin{theorem}
\label{presheaf-theorem}
In the functor categoroid {\bf Set}$^{\cal C}$ over any categoroid ${\cal C}$, every object $P$ is the colimit of a diagram of representable objects, in a canonical way. 
\end{theorem}

{\bf Proof:} This theorem is just the same as the one for (co)presheaves for regular categories by \citet{maclane:sheaves} as Proposition 1 on page 41, Section 1.5 on colimits in functor categories in their book {\em Sheaves in Geometry and Logic}.  The theorem essentially states that given any functoroid $P$ in the categoroid of copresheaves, there is a {\em canonical way} of constructing a small ``index" category ${\cal J}$ and a diagram (which is a functoroid) $A: {\cal J} \rightarrow {\cal C}$ of type ${\cal J}$ such that $P$ is isomorphic to the colimit of the composed diagram ${\cal J} \xrightarrow[]{A} {\cal C} \xrightarrow[]{y}$ {\bf Set}$^{\cal C}$, where the second mapping is the one given by the Yoneda embedding for categoroids. The extension to categoroids involves ensuring that all three types of arrows are mapped over, as the Yoneda Lemma for categoroids already accomplishes using the augmented notion of categoroid natural transformation. 

The strategy behind proving this theorem is to show that the category of copresheaves is Cartesian closed, which have very attractive properties. A category is Cartesian closed if it has a terminal object {\bf 1}, it allows binary products of any two objects $X$ and $Y$, as well as exponentiation $Y^X$. For example, the category {\bf Set} is cartesian closed since clearly it has a terminal object {\bf 1} (the set with 1 object $ = \{ \bullet \}$, and one identity morphism), it allows taking Cartesian products, and it has function objects that are represented  by functions on sets. For the category of copresheaves, showing the first two of these three properties is quite simple. The third, exponentiation, is where the proof is non-trivial, and the reader is referred to \citep{maclane:sheaves} for the last property. For the first two properties, note that the terminal object {\bf 1} in the copresheaf categoroid $\hat{{\cal C}}$ is simply the unique contravariant functoroid that maps the categoroid ${\cal C}^{op}$ to the terminal object in {\bf Set}, that is every object $c$ in ${\cal C}$ is mapped to the one element set, and every morphism from $X$ to $Y$ in ${\cal C}$ is also mapped to the single morphism in {\bf 1}. For products, note that the product of two objects $P$ and $Q$ in the copresheaf categoroid $\hat{{\cal C}}$ is given by the pointwise product of $P(X)$ and $Q(X)$ over every object $X$ in ${\cal C}$. $\bullet$. 

\begin{theorem} 
\label{uct}
{\bf Universal Representation Theorem:} Given any categoroid ${\cal C}$, its copresheaves can be represented as a co-limit of a diagram of representable objects in a unique way.
\end{theorem}

{\bf Proof:} The proof of this theorem builds directly on Theorem~\ref{presheaf-theorem}, which as we just stated is a basic result in the category theory of  copresheaves for regular categories (see \citep{maclane:71}). This result states that {\em any} object $P$ in the functoroid category $\hat{{\cal C}} = $ {\bf Set}$^{\cal C}$ is representable uniquely as a co-limit of a diagram of representable objects. Of course, we already know from the Yoneda Lemma for categoroids that the set of all types of morphisms {\bf Hom}$_{\cal C}(X, -)$ out of an object $X$ is an element of this functoroid category. Given any covariant functoroid $F: {\cal C}  \rightarrow $ {\bf Set}, there is a canonical way to construct a small index diagram $J$  such that  $P$ is isomorphic to the co-limit of the diagram $F: {\cal J} \rightarrow {\cal C} \rightarrow$ {\bf Set}$^{\cal C}$ obtained exactly by the Yoneda Lemma defined above $\bullet$

We now describe a second key result, which states that any an object $X$ in a categoroid can be represented as a natural transformation (a morphism) between two functoroid objects in the presheaf categoroid $\hat{{\cal C}}$. 

\begin{theorem}
\label{crp}
The set of all morphisms  between any two objects $X$ and $Y$ in a categoroid is isomorphic to the set of all natural transformations between the copresheaves defined by $X$ and $Y$, that is {\bf Hom}$_{\cal C}(X,Y) \simeq$ {\bf Nat}({\bf Hom}$_{\cal C}(X, -)$,{\bf Hom}$_{\cal C}(Y, -))$.  
\end{theorem}

{\bf Proof:} The proof of this theorem is a direct consequence of the Yoneda Lemma for categoroids, which states that for every copresheaf functoroid object $F$ in  $\hat{{\cal C}}$ of a categoroid ${\cal C}$, {\bf Nat}({\bf Hom}$_{\cal C}(X, -), F) \simeq F X$. That is, elements of the set $F X$ are in $1-1$ bijection with natural transformations from the presheaf {\bf Hom}$_{\cal C}(X, -)$ to $F$. For the special case where the functoroid object $F = $ {\bf Hom}$_{\cal C}(Y, -)$, we get the result immediately that  {\bf Hom}$_{\cal C}(X,Y) \simeq$ {\bf Nat}({\bf Hom}$_{\cal C}(X, -)$,{\bf Hom}$_{\cal C}(Y,-))$. $\bullet$

\begin{theorem} 
\label{uct-kan}
Any copresheaf in a categoroid is representable as a left Kan extension of a diagram of representable objects in a unique way.
\end{theorem}

{\bf Proof:} Recall from the proof of Theorem~\ref{uct} that {\em any} object $P$ in the functoroid category $\hat{{\cal C}} = $ {\bf Set}$^{\cal C}$ is representable uniquely as a co-limit of a diagram of representable objects. Given any covariant functor $F: {\cal C} \rightarrow $ {\bf Set}, there is a canonical way to construct a small index diagram $J$  that is isomorphic to the co-limit of the diagram $F: {\cal J} \rightarrow {\cal C} \rightarrow$ {\bf Set}$^{\cal C}$ obtained exactly by the Yoneda Lemma for categoroids.  Now, the main step in the reformulation  using the Kan extension is to show that the co-limit of diagram $F$ in the original proof can be itself constructed as a Kan extension. This follows from a more general result that co-limits themselves are just left Kan extension of a functor $F: {\cal C} \rightarrow {\cal E}$ along the functor $L: {\cal C} \rightarrow$ {\bf 1}, where {\bf 1} is the terminal object in category ${\cal C}$. In this application, the functor being extended $F: {\cal J} \rightarrow {\cal C}$ is the diagram functor, as shown below. 

\begin{center}
\begin{tikzcd}[row sep=2cm, column sep=2cm]
% drawing 0- and 1-celss
\mathcal{J}  \ar[dr, "K"', ""{name=K}]
            \ar[rr, "F", ""{name=J, below, near start, bend right}]&&
\mathcal{C}\\
& {\bf 1}    \ar[ur, bend left, "\text{Lan}_KF", ""{name=Lan, below}]
                \ar[ur, bend right, "G"', ""{name=G}]
%
% drawing 2-cells  
\arrow[Rightarrow, "\exists!", from=Lan, to=G]
\arrow[Rightarrow, from=J, to=K, "\eta"]
\end{tikzcd}
\end{center}

Note that the left Kan extension $\mbox{Lan}_K F:$ {\bf 1} $\rightarrow {\cal C}$ essentially has to select a object $c$ of ${\cal C}$, as {\bf 1} contains only one object and one identity mapping. Thus, composing $\mbox{Lan}_K F:$ {\bf 1}  $\rightarrow {\cal C}$ with the functor $K: {\cal J} \rightarrow$ {\bf 1} induces a constant functor $\Delta c$ from the diagram categoroid ${\cal J}$ to the actual categoroid ${\cal C}$, where every object of ${\cal J}$ is mapped to object $c$, and every type of morphism $f: X \rightarrow Y$ in ${\cal J}$ is mapped to the unit morphism $1_c$ of $c$. The universality of the left Kan extension states that {\em every} other functoroid $G$ mapping the unit category {\bf 1} to ${\cal C}$ must factor through the left Kan extension in a unique way. Further, the categoroid natural transformation $\eta$ now maps every object and every morphism in ${\cal F}$ to the constant functor $\Delta_c$. Thus, the constant functor becomes a co-cone under the diagram functoroid, and universality guarantees that this co-cone is indeed a co-limit. $\bullet$

\section{Adjunctions over Categoroids} 

We now define adjunctions over categoroids, which gives us a fundamental tool in analyzing the representations of categoroids using both graphical \citep{pearl:bnets-book} and non-graphical \citep{studeny2010probabilistic} approaches. 

\begin{definition}
\label{adj}
Let ${\cal C}$ and ${\cal C'}$ be two categoroids. A categoroid adjunction between ${\cal C}$ and ${\cal C'}$ is a pair of functoroids $L: {\cal C} \rightarrow {\cal C'}$ and $R: {\cal C'} \rightarrow {\cal C}$ such that for each pair of objects $c$ of ${\cal C}$ and $c'$ of ${\cal C'}$, there is a bijection of sets: 
\begin{equation}
\psi_{{\cal C}, {\cal C'}}: {\cal C'}(L(c), c') \cong {\cal C}(c, R(c'))
\end{equation}
which is natural in both $c$ and $c'$. 
\end{definition}

The functoroid $L$ is then considered a {\em left-adjoint} to $R$, and correspondingly, $R$ is considered a {\em right-adjoint} to $L$. The pair $(L,R)$ is called an adjoint pair of functoroids.  Note that since we are dealing with adjunctions over categoroids, there is a bijection between each of the three types of arrows, so that we get the following set of isomorphisms between corresponding arrows: 

\begin{itemize}
    \item {\bf Regular arrows:} $L(c) \xrightarrow[]{f} c'$ \ \ \ $\cong$ \ \ \ $c \xrightarrow[]{f^{\dagger}} R(c')$ 
    
      \item {\bf Trigonoidal arrows:} $L(a,b,c) \xrightarrow[]{f} (a',b',c) $ \ \ \ $\cong$ \ \ \ $(a,b,c) \xrightarrow[]{f^{\dagger}} R(a', b', c')$ 
      
        \item {\bf Bridge  arrows:} $L((a,b,c)) \xrightarrow[]{f} (a',b') $ \ \ \ $\cong$ \ \ \ $(a, b, c)\xrightarrow[]{f^{\dagger}} R(a',b')$ 
        
           \item {\bf Bridge  arrows:} $L((a,b)) \xrightarrow[]{f} (a',b', c') $ \ \ \ $\cong$ \ \ \ $(a, b)\xrightarrow[]{f^{\dagger}} R(a',b',c)$ 
\end{itemize}

To clarify the notation above, note that for each set of isomorphism of arrows, the collection of arrows on the left are defined on the categoroid ${\cal C}$, whereas the collection of arrows on the right are defined on the categoroid ${\cal C'}$. Furthermore, each arrow $f$ on the left can be viewed as a generalized ``transpose" of the $f^{\dagger}$ arrow on the right. Finally, note that for bridge arrows, as before, the polarity flips between arrows of type $B^0$ and $B^1$ on either side of the adjunction. 

\subsection{Adjunctions between Categoroids and Graphical Models} 

We now illustrate the application of of adjunctions as defined above to the representation of conditional independences by graphical models, a topic that has received a great deal of attention in the past several decades \citep{pearl:bnets-book,lauritzen:chain,mdag,hedge,sadeghi}.  In general, we say a probability distribution $P$ supports a given set of conditional independence relationships, parameterized either as graphoids, imsets, or separoids as defined above. Each distribution $P$ is then represented as an algebraic structure of some kind, for example a directed, undirected, or mixed type of graph. The fundamental question that arises is what dependencies are faithfully captured.  The relationships between what each model can represent and capture is quite complex. DAGs, undirected models, and mixed models capture overlapping sets of relationships, and these are extensively studied in the literature. We turn to discuss a few select examples to give the flavor of the type of result that is known. \citet{sadeghi} gives a detailed discussion of augmented graph structures including directed, undirected, and bidirected edges and the independences that can be represented using them. \citep{studeny2010probabilistic} gives several examples of non-representable conditional independencies. 

In terms of our  definition of adjunctions, if we define an adjunction between two functoroids $F: {\cal C} \rightarrow {\cal D}$, where ${\cal D}$ is the categoroid of all DAG models, and its right adjoint $G: {\cal D} \rightarrow {\cal C}$, it is clear that the functoroid ${\cal F}$ is not surjective. Indeed, \citep{studeny2010probabilistic} notes that over just $4$ variables, there are over $18,000$ conditional independent structures, but DAG models can only faithfully render a few hundred of these. The most we can then hope for is that this functoroid mapping is injective, that is $1-1$. Unfortunately, even that fails to be the case.  Non-identifiability of DAG models, such as the serial model $A \rightarrow B \rightarrow C$, the diverging model $B \leftarrow A \rightarrow C$ and the reverse serial model $C \rightarrow B \rightarrow A$, occurs because given the parameterization of one of the models, the parameterization of the others can be derived using Bayes rule.

\subsection{Free Objects in Categoroids} 

We can define more formally the notion of a ``free object" in a categoroid, in terms of adjoint functoroids. Given a pair of adjoint functoroids $L, R$ as defined in Definition~\ref{adj}, if we consider $R = U$ to be a ``forgetful" functoroid that ``throws away" structure, and its left adjoint $L$ exists, then the defining property of an adjunction means that for each arrow (of any type) from object $c$ to $U(c')$ in the underlying categoroid, there is a unique corresponding morphism (of the corresponding type) from $L(c)$ to $c'$, which defines $L(c)$ as the ``free object" associated with $c$. 

To instantiate this notion of free object in the context of the previous discussion, let us define $R$ as the functoroid from the category of graphical models, say DAGs, to the underlying categoroid representing the conditional independence relations, and $L$ to be its left adjoint in the reverse direction. We can now interpret the three DAGs that all define the conditional independence property $A \CI C | B$ as the ``free objects" associated with the underlying categoroid, whereas the collider is the free object associated with the categoroid defined by $A \CI C | \emptyset$. Adjunctions give a rigorous framework to analyze the faithfulness of both graphical and non-graphical representations of categoroids. In the next section, we introduce a powerful way to construct representations of any categoroids using left {\bf k}-modules over a ring. 

\subsection{Integer-Valued Multisets as Adjunctions} 

We can now revisit the idea described above of how integer-valued multisets essentially form a modular representation of a categoroid, which defines a right-adjoint between a imset, viewed as an object in the category {\bf Set}, and the free object defined by the Abelian group representing the formal linear sums with integer coefficients of the free group defined by the set. Recall an imset  defines an integer-valued multiset function $u: \mathbb{Z}^{{\cal P(\mathbb{Z})}} \rightarrow \mathbb{Z}$ from the power set of integers, ${\cal P(\mathbb{Z})}$ to integers $\mathbb{Z}$, where a {\em combinatorial} imset is defined as: 

\[ u = \sum_{A \subset N} c_A \delta_A \]

where $c_A$ is an integer, and $A$ potentially ranges over all subsets of $N$. An {\em elementary} imset is defined over $(a,b \CI A)$, where $a,b$ are singletons, and $A \subset N \setminus \{a,b\}$. A {\em structural} imset is defined as one where the coefficients can be rational numbers. Imsets can be viewed as a modular representation of a categoroid in terms of adjunctions defined by a pair of functors between the category of Abelian groups {\bf Ab} and the category of sets {\bf Set} \citep{riehl2017category}. The ``free" Abelian group on a given set $S$ is defined formally as 

\[ \mathbb{Z}[S] = \sum_{x \in S} z_x x \] 

where $z_x$ is the integer coefficient associated with the element $x \in X$. In the case of imsets, the objects are defined as the triples, such as $(ac|b)$ as shown in DAG models in Figure~\ref{imset}. A ``forgetful" functor $U$ defines the {\em right-adjoint} between the Abelian group $\mathbb{Z}[S]$ and the set $S$ (where the group structure is ``thrown away" in the category of sets {\bf S}, of which $S$ is an object). The ``free" functor $L$ maps the set $S$ back to the Abelian group $\mathbb{Z}[S]$, Thus, imsets are a particular example of a much broader class of adjoint functors defined by ``free" $+$ ``forgetful" functors, which are widely used in category theory

\section{Summary and Future Work} 

We proposed categoroids, a type of category, to study universal properties of conditional independence. We related categoroids to three axiom systems -- graphoids, integer-valued multisets, and separoids -- showing how each of these can be interpreted as a categoroid.  We suitably modified the standard definitions of cateogry theory, including functors and natural transformations, to capture the binary and ternary structure. Specifically, we defined functoroids to map from one categoroid to another, preserving not just the binary  structure defining by the algebraic structure, but also the ternary relations capturing the triples of elements satisfy the conditional independence axioms. Natural transformations map across two functoroids. We proved a variant of the Yoneda Lemma to construct a universal representation of a conditional independence system in an abstract category. Our framework leads to new insights across the previous axiom systems, such as recognizing that integer multisets can be viewed as inducing morphisms through M\"obius inversions, and the mapping from a conditional independence system to a graphical or non-graphical representation can be viewed in terms of adjunctions involving free and forgetful functors. We focused on the structure of presheaves and copresheaves as the principal carriers of conditional independence properties. Much remains to be done in this paradigm, and we describe a few extensions of this work, some of which are already in progress: 

\subsection{Modular Representations of Categoroids} 

\begin{figure}[h] 
\centering
\caption{Modular representations over categoroids can be defined using the Yoneda Lemma to construct a module over the co-presheaf of all three types of morphisms exiting an object. \label{ci-quiver}}
\vskip 0.1in
\begin{minipage}{0.6\textwidth}
\vskip 0.1in
\includegraphics[scale=0.3]{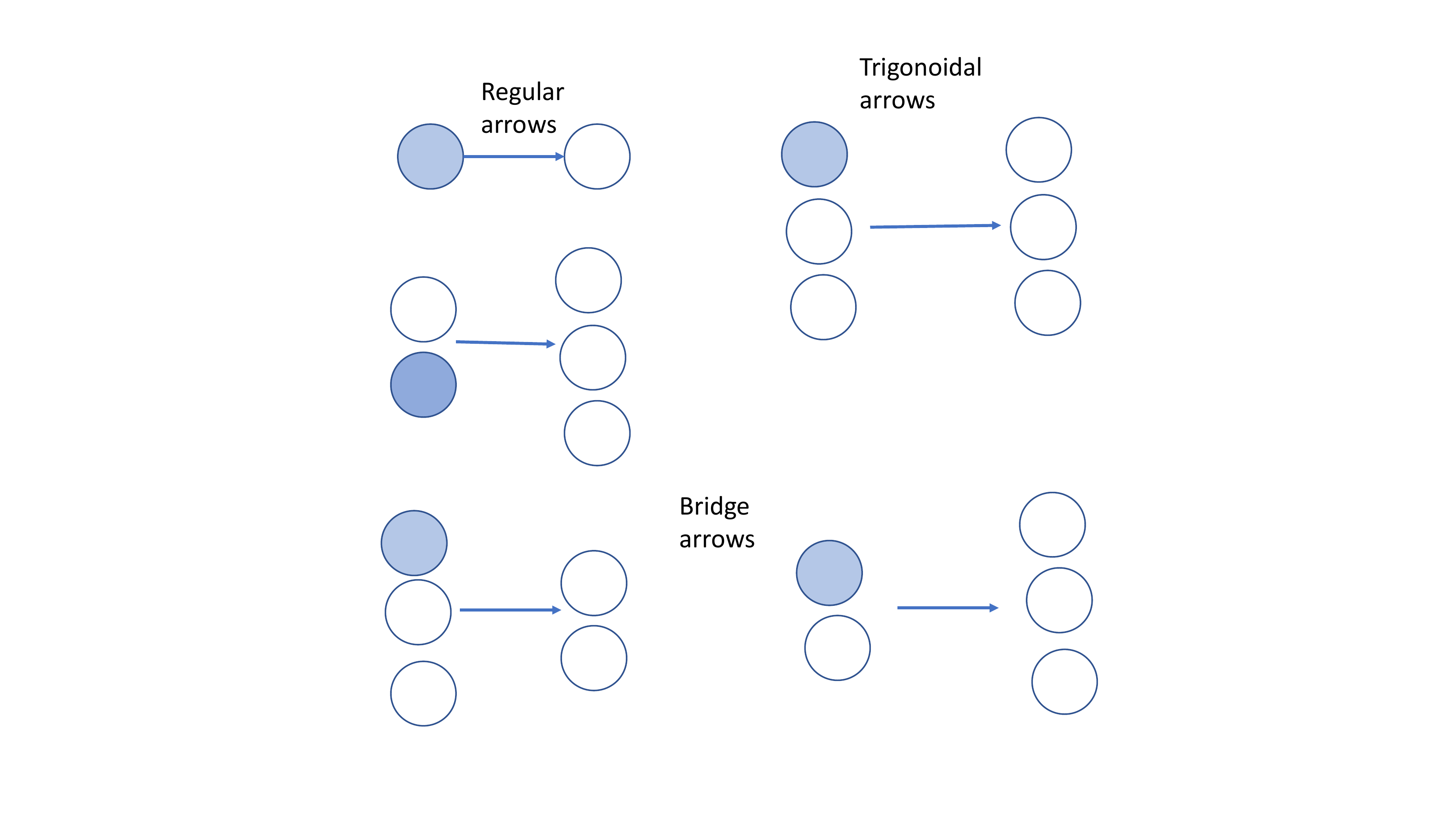}
\end{minipage}
\end{figure}

One direction for future work is to construct Gr\"obner representations of categoroids. Recent work has defined Gr\"obner representations for combinatorial categories \cite{Sam_2016}. Broadly speaking,  this approach  generalizes the work on modeling graphical models as algebraic varieties \citep{cbn,BEERENWINKEL2006409,Geiger_2006}, and ideals on partially ordered sets (posets) \citep{hibi1987distributive}. The intuitive idea is that a representation of a categoroid can be defined as an abstract Gr\"obner basis over an ideal defined on a module whose basis is defined using the free algebra generated by the set of all morphisms out of an object (including all three arrow types). \citet{Sam_2016} define a general {\bf Gr\"obner} category as one that satisfies certain conditions on the structure of posets, which enables the results to go through. 

Let $X$ be a (possibly non-finite) poset $X$ satisfies the {\bf ascending chain condition} (alternatively called well ordering) if every ascending chain in $X$ stabilizes, that is, given $x_1 \leq x_2 \leq \ldots $ in $X$, $x_i \nleq x_j$, we have $x_i = x_{i+1}$ for $i \gg 0$. The {\bf descending chain condition} is defined analogously. An {\bf anti-chain} in $X$ is any set of incomparable elements, that is, for any sequence $x_1, x_2, \ldots$, any pair $x_i \nleq x_j$ for all $i \neq j$. An {\bf ideal} in a poset $X$ is a subset $I$ that is downwards (or upwards) closed, that is if $x \in I$, and $x \leq y$, then $y \in I$ (analogous definition for the complementary relation). Denote ${\cal I}(X)$ for the poset of ideals, ordered by subset containment. For any $x \in X$, the {\bf principal ideal} generated by $x$ is $\{y | y \geq x\}$. An ideal is {\bf finitely generated} if it a finite union of principal ideals. If $X$ and $Y$ are noetherian posets, their product $X \times Y$ is noetherian as well. $f: X \rightarrow Y$ is an {\bf order-preserving} function between two posets $X$ and $Y$ if $x \leq x' \Rightarrow f(x) \leq f(x')$. 

Specifically, denote {\bf Rep}$_{\bf k}({\cal C})$ as the category of representations of a categoroid ${\cal C}$, where {\bf k} is a non-zero ring, and Mod$_{\bf k}$ is the category of left-{\bf k} modules. Thus, we can define a {\bf representation} of a categoroid ${\cal C}$ as a functoroid ${\cal C} \rightarrow \mbox{Mod}_{\bf k}$.  Let $x$ be an object of a categoroid ${\cal C}$, e.g., a normal object or a trigonoidal object.  Define a representation $P_x$ of ${\cal C}$ as a left {\bf k}-module, where $P_x(y) = {\bf k}[{\bf Hom}_{\cal C}(x,y)]$, that is, $P_x(y)$ is the free left {\bf k}-module with basis {\bf Hom}$_{\cal C}(x,y)$. For any particular morphism $f: x \rightarrow y$, let $e_f$ denote the corresponding element of $P_x(y)$.  We can extend the approach developed by \citet{Sam_2016} to include bridge and trigonoidal morphisms as well. If $M$ is any other representation, we can construct a homomorphism {\bf Hom}$(P_x,M) = M(x)$. Thus, it can be shown that {\bf Hom}$(P_x, -)$ is an exact functor, and thus $P_x$ is a projective object of {\bf Rep}$_{\bf k}$. An object of {\bf Rep}$_{\bf k}$ is defined to be {\bf noetherian} if every ascending chain of subobjects stabilizes. The category {\bf Rep}$_{\bf k({\cal C})}$ is noetherian if every finitely generated object in it is. We refer the reader to \citep{Sam_2016} for further background and detailed proofs. 

\subsection{Universal Constructions in Categories} 

Category theory provides a rich set of construction tools that facilitate modeling complex types of interactions (see Figure~\ref{operads}.  Many applications of category theory use monoidal categories based on defining a tensor product operator $\otimes: {\cal C} \times {\cal C} \rightarrow {\cal C}$. Monoidal categories are incredibly useful in a variety of rich applications, as  detailed in the book by \citet{fong2018seven}.  We can define categoroids using construction techniques, such as decorated cospans, hypergraph categories, and operads, to mention a few of the many choices available. These enhancements will be explored in subsequent paper. 

\begin{figure}[t]
\begin{center}
\begin{small}
\begin{tabular}{|c |c | } \hline 
{\bf Construction Tool } & {\bf Application} \\ \hline 
Braiding and tensor products \citep{JOYAL1996164} & Control theory \citep{Baez_2010} \\ \hline 
 Galois Extensions and Profunctors \citep{fong2018seven}& Engineering design \citep{censi} \\ \hline 
Co-limits and Decorated cospans \citep{cospans}  & Electric Circuits \citep{fong2018seven} \\ \hline
Operads \citep{higher-operads} & Databases \citep{operads-db}  \\ \hline
Hypergraph categories \citep{fong2018seven} & Signal flow diagrams \\ \hline 
Bisimulation morphisms \citep{DBLP:conf/lics/JoyalNW93} & Concurrent systems \citep{bisim} \\ \hline
\end{tabular}
\end{small} 
\end{center}
\caption{Category theory offers a rich repertoire of construction tools to build categoroids.}
\label{operads}
\end{figure} 

\subsection{Topoids: Categoroids over Toposes} 

We can define a large variety of categoroids using specific types of well-known categories. In this section, we give one example, using the category of toposes (or topoi) \citep{goldblatt:topos,Johnstone:592033,maclane:sheaves}. 

\begin{definition}
\label{topoid}
A {\bf topoid} ${\cal C}$ is a categoroid, where the underlying category is a {\em topos}. That is, we assume the underlying category has all co-limits and limits, has exponential objects, and a subobject classifier $1 \rightarrow \Omega$. To be a topoid, the topos has to be augmented with three types of arrows, as in a categoroid, representing both the usual binary category structure and the ternary conditional independence structure, along with the brige morphisms defining the join. 
\end{definition}

In effect, toposes \citep{goldblatt:topos,maclane:sheaves} are categories that admit all the nice properties of sets. Recall in sets that we can always define the product of two sets $A \times B$ (the limit), the (disjoint) union of two sets $A \sqcup B$ (the colimit), we can always define functions $f: A \rightarrow B$, between two sets, which can be viewed as exponential objects in the set of all functions $B^A$, and we can define subsets $C \subset A$, whose subobject classifier is the characteristic function $\chi_C: A \rightarrow \{0,1 \}$, where $\chi_C(x) = 1$ if and only if $x \in C$. Toposes abstract out each of these properties and use it to define a more abstract notion of sets, which has found great utility in unifying many areas of mathematics from logic to geometry and topology. 

\subsection{Monads: Monoids in the Categoroid of Endofunctoroids} 

Monads are one of the most widely used constructions used in category theory to gain insight of a ``target" category from a ``source" category. Recall from the definition of adjunctions in a categoroid that we are given two functoroids, $L: {\cal C} \rightarrow {\cal C'}$, and $R: {\cal C'} \rightarrow {\cal C}$,  which operate in opposing directions. From the standpoint of categoroid ${\cal C}$, what insight can we gain about categoroid ${\cal C'}$ if we cannot ``observe" it directly, but only observe it indirectly through the composite functoroid $R \circ F: {\cal C} \rightarrow {\cal C}$. Such a composite functoroid defines an {\em endofunctoroid}, namely a mapping from a categoroid ${\cal C}$ to itself. The set of all endofunctoroids from a categoroid to itself forms a categoroid. The {\em monad} on this categoroid of endofunctoroids is a special object, which we can illustrate with a simpler example of a monoid on a set. 

\begin{definition}
\label{monad} 
A {\bf monoid} is an object $M$ in the category {\bf Set}, together with a pair of morphisms $\mu: M \times M \rightarrow M$ and $\eta: I \rightarrow M$ such that the following diagram commutes: 
\begin{center}
\begin{tikzcd}
M \times M \times M \arrow[r, "1_\mu \times \mu"] \arrow[d, "\mu \times 1_M", red]
& M \times M \arrow[d, "\mu" red] \\
M \times M \arrow[r, red, "\mu" blue]
& |[blue]| M 
\end{tikzcd}
\end{center} 
\end{definition}

We aim to conduct a deeper study of conditional axiom systems through the use of algebras defined over monads \citep{riehl2017category}.

\subsection{Computational Issues} 

We have ignored the crucial aspect of computational tractability in this paper, as our work is primarily  focused on understanding the deeper algebraic properties of conditional independences, generalizing previous formalisms such as separoids \citep{DBLP:journals/amai/Dawid01} and imsets \cite{studeny2010probabilistic}. We have previously studied the use of homotopy to define equivalences among causal models \citep{sm:homotopy}, showing a more efficient enumeration of possible DAGs using homotopic equivalences. The use of homotopy in category theory is a rich area that allows many interesting ways of finding compact representations \citep{richter2020categories}, a direction we intend to study in the future. 

%\bibliographystyle{unsrtnat}  
%bibliography{references,allcitations}

\end{document}